\documentclass[twoside]{article}

\usepackage[accepted]{aistats2022}

\usepackage{algorithm}
\usepackage{algorithmic}
\usepackage{latexsym, amsmath, amssymb, amsthm}
\usepackage{graphicx}
\usepackage{subfig}
\usepackage[algo2e, ruled, vlined, linesnumbered]{algorithm2e}
\usepackage{wrapfig}
\usepackage{adjustbox}

\usepackage[round]{natbib}

\usepackage{appendix}

\usepackage[colorlinks=true]{hyperref}
\usepackage[dvipsnames]{xcolor}
\hypersetup{
    allcolors = MidnightBlue,
}

\DeclareMathOperator*{\argmax}{arg\,max}
\newcommand{\x}{\mathbf{x}}
\newcommand{\z}{\mathbf{z}}
\newcommand{\prob}[1]{\mathbb{P}\left[{#1}\right]}
\newcommand{\expec}[2]{\mathbb{E}_{#2}\left[#1\right]}
\newcommand{\kl}[2]{\text{KL}\left(#1||#2\right)}
\newcommand{\bs}[1]{\boldsymbol{#1}}
\newcommand{\pert}{\bs{\delta}}
\DeclareMathOperator*{\argmin}{arg\,min}
\newcommand{\pertsample}{\z_{\pert}}
\newcommand{\unpertsample}{\z_{\neg \pert}}
\newcommand{\diag}[1]{\texttt{diag}\left(#1\right)}
\newcommand{\gauss}[3]{\mathcal{N}\left(#1; #2, #3\right)}

\let\oldnl\nl
\newcommand{\nonl}{\renewcommand{\nl}{\let\nl\oldnl}}

\newtheorem{theorem}{Theorem}

\newtheorem{lemma}[theorem]{Lemma}

\theoremstyle{definition}
\newtheorem{definition}{Definition}[section]

\begin{document}

\runningauthor{Barrett, Camuto, Willetts, Rainforth}

\twocolumn[

\aistatstitle{Certifiably Robust Variational Autoencoders}

\aistatsauthor{ Ben Barrett$^{1}$ \And Alexander Camuto$^{1, 3}$ \And  Matthew Willetts$^{2, 3}$ \And Tom Rainforth$^{1}$ }

\aistatsaddress{ $^{1}$University of Oxford \And $^{2}$University College London \And $^{3}$Alan Turing Institute} ]

\begin{abstract}
  We introduce an approach for training variational autoencoders (VAEs) that are certifiably robust to adversarial attack.
  Specifically, we first derive actionable bounds on the minimal size of an input perturbation required to change a VAE's reconstruction by more than an allowed amount, with these bounds depending on certain key parameters such as the Lipschitz constants of the encoder and decoder.
  We then show how these parameters can be controlled, thereby providing a mechanism to ensure \textit{a priori} that a VAE will attain a desired level of robustness.
  Moreover, we extend this to a complete practical approach for training such VAEs to ensure our criteria are met.
  Critically, our method allows one to specify a desired level of robustness \emph{upfront} and then train a VAE that is guaranteed to achieve this robustness.
  We further demonstrate that these \emph{Lipschitz--constrained} VAEs are more robust to attack than standard VAEs in practice.
\end{abstract}

\section{INTRODUCTION}

Variational autoencoders (VAEs) are a powerful method for learning deep generative models~\citep{kingma2013autoencoding, rezende2014stochastic}, finding application in areas such as image and language generation \citep{razavi2019generating, kim2018semiamortized} as well as representation learning \citep{Higgins2017betaVAELB}.

Like other deep learning methods \citep{szegedy2013intriguing}, VAEs are susceptible to adversarial attacks, whereby small perturbations of an input can induce meaningful, unwanted changes in output.
For example, VAEs can be induced to reconstruct images similar to an adversary's target through only moderate perturbation of the input image \citep{tabacof2016adversarial, gondimribeiro2018adversarial, Kos_2018}.

\begin{figure*}[t!]
    \captionsetup[subfloat]{farskip=-1pt,captionskip=-1pt}
    \begin{tabular}{cc}
    \centering
        \subfloat[Standard VAE, $||\pert||_2 \leq 3$.]{\includegraphics[width=0.46\textwidth]{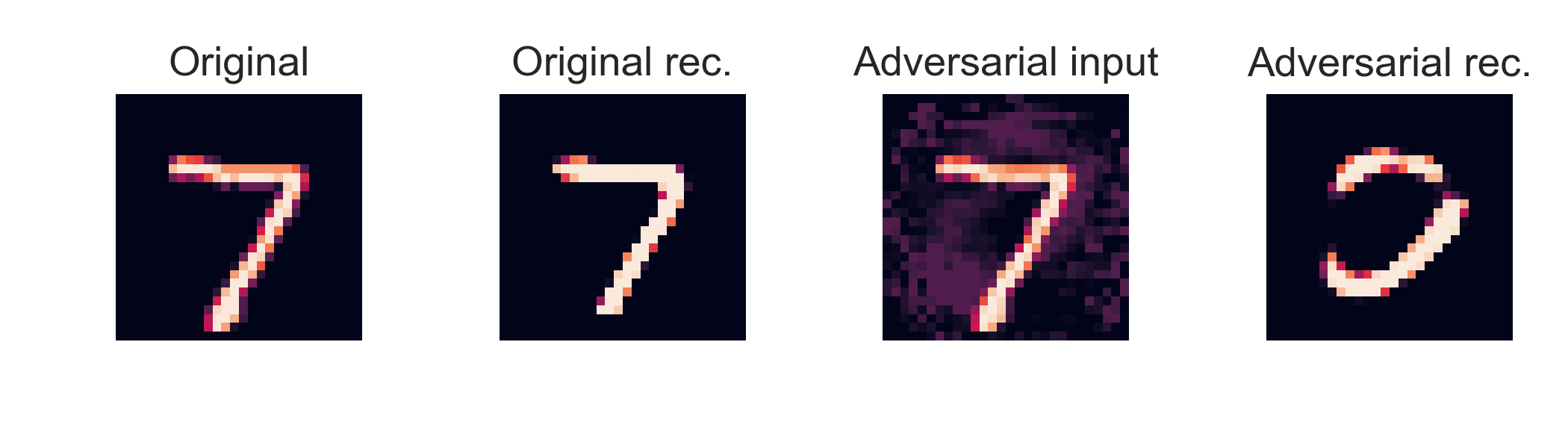}} & \subfloat[Lipschitz-constrained VAE, $||\pert||_2 \leq 3$.]{\includegraphics[width=0.46\textwidth]{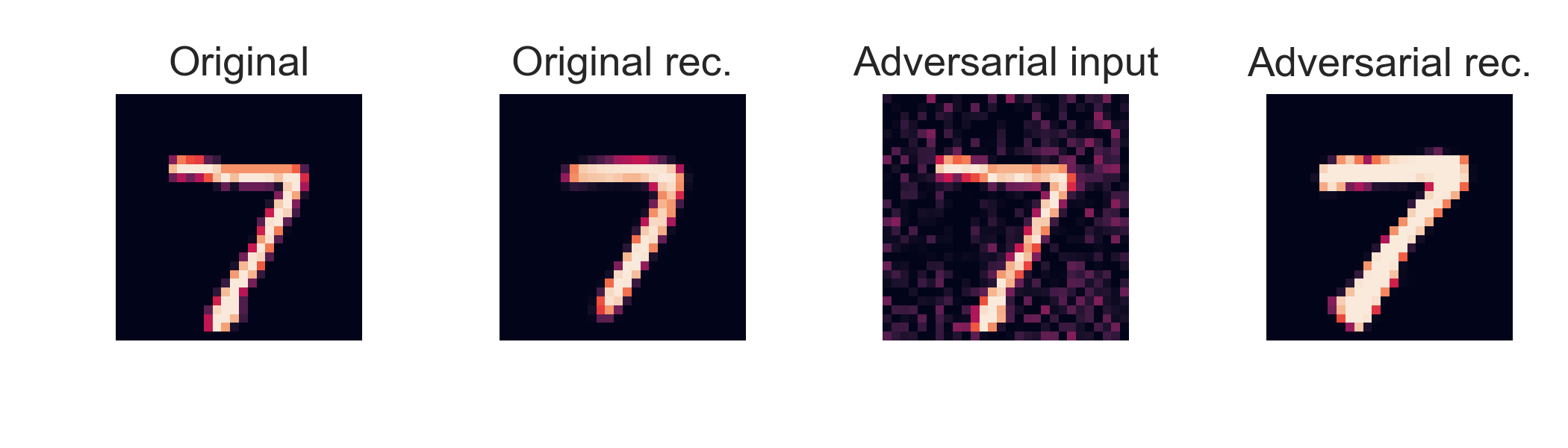}}
    \end{tabular}
    \caption{A maximum damage attack \eqref{eq: max_damage_attack} on a standard VAE and Lipschitz-constrained VAE, respectively, for the same perturbation norm constraint. Unlike those of the standard VAE, the Lipschitz-constrained VAE's reconstructions are robust under attack.
    Appendix \ref{app:qualitative_eval} supplements these results with latent space attacks \eqref{eq: latent_space_attack_2}.
    }
    \label{fig:illustrative_example}
\end{figure*}

Two reasons why this is particularly undesirable are a) that VAEs have been used to improve the robustness of classifiers \citep{schott2018adversarially, Ghosh_2019} and b) the encodings of VAEs are commonly used in downstream tasks \citep{ha2018, higgins2017darla}.
Yet another is that the susceptibility of VAEs to distortion from input perturbations challenges an original ambition for VAEs: that they should capture ``semantically meaningful [...] factors of variation in data'' \citep{Kingma_2019}.
If this ambition is to be fulfilled, VAEs should be more robust to spurious inputs,
and so the robustness of VAEs is intrinsically desirable.

While previous work has already sought to obtain more robust VAEs empirically \citep{willetts2021improving, cemgil2020advtraining, cemgil2020autoencoding},
this work lacks formal guarantees.
This is a meaningful worry because in other model classes, robustification techniques showing promise empirically, but lacking guarantees, have later been circumvented by more sophisticated attacks \citep{athalye2018obfuscated, uesato2018adversarial}.
It stands to reason that existing techniques for robustifying VAEs might be similarly ineffectual.
Further, though previous theoretical work \citep{alex2020theoretical} can ascertain robustness \textit{post-training}, it cannot enforce and control robustness \textit{a priori}, before training.

Our work looks to alleviate these issues by providing VAEs whose robustness levels can be controlled and certified by design.
To this end, we show how \textit{certifiably robust} VAEs can be learned by enforcing Lipschitz continuity in the encoder and decoder, which explicitly upper-bounds changes in their outputs with respect to changes in input.

We derive two bounds on the robustness of these models, each covering a slightly different setting.
First, we derive a per-datapoint lower bound that guarantees a minimum probability for reconstructions of distorted inputs being within some distance of the reconstructions of undistorted inputs.
More precisely, this per-datapoint lower bound is on the probability that the $\ell_2$ distance between
an attacked VAE's reconstruction and its original reconstruction
is less than some value $r$.
This probability is with reference to the stochasticity
of sampling in a VAE's latent space.
Using the previous bound, we can then obtain a margin that holds for all inputs.
This second, \textit{global} bound means that we can guarantee, for \textit{any} input, that perturbations within the margin induce reconstructions that fall within an $r$--sized ball of the original reconstruction with \textit{at least} some specified probability $\epsilon$.

The latter margin is the first of its kind for VAEs: a margin that is input-agnostic and can have its value specified $\textit{a priori}$ from setting a small number of network hyperparameters.
It thus enables VAEs with chosen levels of robustness.

In summary, our key contributions are to a) develop the first \emph{certifiably robust} VAE approach, wherein
Lipschitz continuity constraints are used during training to ensure certain robustness properties are met; b) provide accompanying theory to show that our approach allows a desired level of robustness to be  \emph{guaranteed upfront}; and c) experimentally validate that our approach works in practice (see e.g.~Figure~\ref{fig:illustrative_example}).

\section{BACKGROUND}

\subsection{VAEs}
Assume we have a collection of observations $\mathcal{D}=\left\{\x_1, \ldots, \x_n\right\}$ with $\x \in \mathcal{X}$, which is generated according to an unknown process involving latent variables $\z \in \mathcal{Z}$.
We want to learn a latent variable model with joint density $p_\theta(\x, \z)=p_\theta(\x|\z)p(\z)$, parameterized by $\theta$, that captures this process.
Learning $\theta$ by maximum likelihood is often intractable;
variational inference addresses this intractability by introducing inference model $q_\phi(\z|\x)$ \citep{Kingma_2019}, parameterized by $\phi$, which yields the ``ELBO", a tractable lower bound on the marginal log likelihood $\log p_\theta(\x)$,
\begin{align}
\label{eq: likelihood_lower_bound}
   \expec{\log p_\theta(\x|\z)}{q_\phi(\z|\x)}\text{--}\kl{q_\phi(\z|\x)}{p(\z)}.
\end{align}
Here, $\kl{\cdot}{\cdot}$ denotes the Kullback-Leibler divergence, while
$\theta$ and $\phi$ represent the parameters of neural networks --- the \emph{decoder} and \emph{encoder network}, respectively, of the VAE --- which can be optimized using unbiased gradient estimates obtained through Monte Carlo samples from $q_\phi(\z|\x)$.

Given a VAE, we will refer to sampling $\z_i \sim q_\phi(\z|\x_i)$ on input $\x_i$ as the \emph{encoding process}, and, following convention, to $g_\theta(\z_i)$ as a \emph{reconstruction} of $\x_i$, where $g_\theta(\cdot)$ denotes the \emph{deterministic component of the decoder} \citep{kumar2020implicit}.

\subsection{Adversarial Attacks On VAEs}
\label{sec: adv_attacks_VAEs}
In adversarial attacks, an adversary tries to alter the behavior of a model.
Although much work has focused on classifiers, adversarial attacks have also been proposed for VAEs, whereby the model is ``fooled'' into reconstructing an unintended output.
More formally, given original input $\x_o$ and the adversary's target output $\x_t$, the attacker seeks a perturbation $\pert \in \mathcal{X}$ such that the VAE's reconstruction of perturbed input $(\x_o + \pert)$ is similar to $\x_t$.

The best performing attack on VAEs in the current literature is a \emph{latent space attack} \citep{tabacof2016adversarial, gondimribeiro2018adversarial, Kos_2018}, where an adversary perturbs input $\x_o$ to have a posterior $q_\phi$ similar to that of the target $\x_t$, optimizing
\begin{equation}
\label{eq: latent_space_attack}
\argmin_{\pert} \quad \kl{q_\phi(\z|\x_o + \pert)}{q_\phi(\z|\x_t)} + \lambda ||\pert||_2.
\end{equation}
In \eqref{eq: latent_space_attack}, the second term implicitly constrains the perturbation norm; in our work, we explicitly constrain this norm by some constant $c \in \mathbb{R}^+$ to ensure more consistent comparisons:
\begin{equation}
\label{eq: latent_space_attack_2}
\argmin_{\pert:\ ||\pert||_2\leq c} \quad \kl{q_\phi(\z|\x_o + \pert)}{q_\phi(\z|\x_t)}.
\end{equation}

We also use another type of attack, the \emph{maximum damage attack} \citep{alex2020theoretical}, which for $\pertsample \sim q_\phi(\z|\x_o+\pert)$, $\unpertsample \sim q_\phi(\z|\x_o)$, and some constant $c \in \mathbb{R}^+$ optimizes
\begin{equation}
\label{eq: max_damage_attack}
\argmax_{\pert:\ ||\pert||_2\leq c} \quad ||g_\theta(\pertsample) - g_\theta(\unpertsample)||_2.
\end{equation}

\subsection{Defining Robustness In VAEs}

VAE reconstructions are typically continuous--valued, and a VAE's encoder, $q_\phi(\z|\x)$, is usually chosen to be a continuous distribution.
Any change to a VAE's input will thus almost surely result in a change in its reconstructions, since changes to the input will translate to changes in $q_\phi(\z|\cdot)$ and, in turn, to changes in the reconstruction $g_\theta(\z)$ \citep{alex2020theoretical}.

This observation rules out established robustness criteria that specify robustness using margins around inputs within which model outputs are constant \citep{cohen2019certified, salman2019provably}. To further complicate matters, VAEs are probabilistic: a VAE's outputs will vary even under the same input.
To account for these considerations, we employ the robustness criterion of \cite{alex2020theoretical}:

\theoremstyle{definition}
\begin{definition}{(($r, \epsilon$)-robustness)}
\label{def: r_robust}
For $r\in \mathbb{R}^+$ and $\epsilon \in [0, 1)$, a model $f$ operating on a point $\x$ and outputting a continuous random variable is ($r, \epsilon$)-robust to a perturbation $\pert$ if and only if
\[\prob{||f(\x+\pert)-f(\x)||_2 \leq r} > \epsilon.
\footnote{We use the $\ell_2$ norm but the following definitions could also be stated with respect to other norms.}
\]
\end{definition}
The notion of $(r, \epsilon)$-robustness states that a model is robust if, with probability greater than $\epsilon$, changes in the model's outputs induced by input perturbation $\pert$ fall within a hypersphere of radius $r$ about the model's outputs on the unperturbed input. The smaller the $r$ and the larger the $\epsilon$ for which $(r, \epsilon)$-robustness holds, the stricter the notion of robustness which is implied. We will refer to $\prob{||f(\x+\pert)-f(\x)||_2 \leq r}$ as the \emph{$r$-robustness probability}. Note that, by enabling flexible specification of $r$ and $\epsilon$, the $(r, \epsilon)$-robustness criterion can be made \emph{arbitrarily strong} to suit the level of robustness required.

The notion of $(r, \epsilon)$-robustness naturally yields that of an \emph{$(r, \epsilon)$-robustness margin} \citep{alex2020theoretical}:
\begin{definition}{(($r, \epsilon$)-robustness margin)}
    \label{def: robustness_margin}
    For $r\in \mathbb{R}^+$ and $\epsilon \in [0, 1)$, a model $f$ has ($r, \epsilon$)-robustness margin $R^{(r, \epsilon)}(\x)$ about input $\x$ if
    $||\pert||_2 < R^{(r, \epsilon)}(\x) \implies \prob{||f(\x+\pert)-f(\x)||_2 \leq r} > \epsilon$.
\end{definition}
A model with an $(r, \epsilon)$-robustness margin on $\x$ can only be undermined by more than $r$ by perturbations with norm less than $R^{(r, \epsilon)}(\x)$ with probability less than $(1-\epsilon)$.
In other words, for appropriately chosen $r$ and $\epsilon$, we can guarantee that a model with an $(r, \epsilon)$-robustness margin cannot be consistently undermined by input perturbations of $\x$ up to a particular magnitude \citep{alex2020theoretical}.

\subsection{Lipschitz Continuity}

For completeness, recall the following definition:
\theoremstyle{definition}
\begin{definition}{(Lipschitz continuity)}
\label{def: lipschitz_continuity}
A function $f: \mathbb{R}^n \rightarrow \mathbb{R}^m$ is Lipschitz continuous if for all $\x_1, \x_2 \in \mathbb{R}^n$,
$||f(\x_1)-f(\x_2)||_2 \leq M||\x_1-\x_2||_2$
for constant $M \in \mathbb{R}^+$.
The least $M$ for which this holds is called the \emph{Lipschitz constant} of $f$.
\end{definition}

If a function $f$ is Lipschitz continuous with Lipschitz constant $M$, we say $f$ is $M$-Lipschitz.

\section{CERTIFIABLY ROBUST VAES}
\label{sec: theory}

We now introduce our approach for achieving a VAE whose robustness levels can be controlled and certified.
We do so by targeting the ``smoothness'' of a VAE's encoder and decoder network, requiring these to be Lipschitz continuous,
since a VAE's vulnerability to input perturbation is thought to inversely correlate with the smoothness of its encoder and decoder.
By maintaining Lipschitz continuity with known, set Lipschitz constants, we will be able to obtain a chosen degree of robustness \textit{a priori}.

\subsection{Bounding The $r$-Robustness Probability}
We first construct an approach for guaranteeing that a VAE's reconstructions will change only to a particular degree under input distortions.
We achieve this by specifying our VAEs such that their $r$-robustness probability is bounded from below.

In the standard setting, this yields an input-dependent characterization of the behavior of the VAE, while bounding the encoder standard deviation or taking it to be a hyperparameter yields global, input-agnostic bounds.
This means that for a given input perturbation norm we can guarantee output similarity up to a threshold with a particular probability.
Our bounds provide the first global guarantees about the robustness behavior of a VAE.

We use the $\ell_2$ distance as our notion of similarity because it has been the basis for previous theoretical work on VAE robustness~\citep{alex2020theoretical} and also corresponds to the log probability of a Gaussian --- a frequently-used likelihood function for VAEs.

The following result shows that, under the common choice of a diagonal-covariance multivariate Gaussian encoder, a lower bound on the $r$-robustness probability can be provided for VAEs.\footnote{Note that our results operate on the basis that the forward pass involves sampling both at train and test time; sampling at test time is not unreasonable because of the adversarial setting, wherein a VAE's sampling step may be critical to the robustness of a downstream task.} We use the parameterization
$q_\phi(\z|\x)=\gauss{\z}{\mu_\phi(\x)}{\diag{\sigma_\phi^2(\x)}}$,
where %
$\mu_\phi: \mathcal{X} \rightarrow \mathbb{R}^{d_z}$ is the \emph{encoder mean}
and $\sigma_\phi: \mathcal{X} \rightarrow \mathbb{R}^{d_z}_{\geq 0}$ is the \emph{encoder standard deviation}.\footnote{We assume $\mathcal{X}$ to be Euclidean.}

\begin{theorem}[Probability Bound]
\label{thm: probability_bound}
Assume $q_\phi(\z|\x)$ is as above
and that the deterministic component of the VAE decoder $g_\theta(\cdot)$ is $a$-Lipschitz, the encoder mean $\mu_\phi(\cdot)$ is $b$-Lipschitz, and the encoder standard deviation $\sigma_\phi(\cdot)$ is $c$-Lipschitz. Finally, let $\pertsample \sim q_\phi(\z|\x+\pert)$ and $\unpertsample \sim q_\phi(\z|\x)$. Then for any $r\in \mathbb{R}^+$, any $\x \in \mathcal{X}$, and any input perturbation $\pert \in \mathcal{X}$,
\[\prob{||g_\theta(\pertsample) - g_\theta(\unpertsample)||_2 \leq r} \geq 1 - \min\left\{p_1(\x), p_2(\x) \right\},\]
where
\begin{align*}
p_1(\x)\!&:=\!\min\left(1, \frac{a^2\left(b^2||\pert||_2^2 + (c||\pert||_2+2||\sigma_\phi(\x)||_2)^2\right)}{r^2}\right) \\
p_2(\x) & := \begin{cases}
C(d_z) \frac{u(\x)^{\frac{{d_z}}{2}}\exp\left\{-\frac{u(\x)}{2}\right\}}{u(\x)-{d_z}+2}  & \left(\frac{r}{a}-b||\pert||_2\right) \geq 0; \\
& {d_z} \geq 2;\\
& u(\x) > {d_z}-2\\
1 & \text{o.w.}
\end{cases}
\end{align*}
for
$u(\x) := \frac{\left(\frac{r}{a}-b||\pert||_2\right)^2}{\left(c||\pert||_2 + 2||\sigma_\phi(\x)||_2\right)^2}$ and constant $C({d_z}) := \frac{1}{\sqrt{\pi}}\exp\left\{\frac{1}{2}({d_z}-({d_z}-1)\log {d_z})\right\}.$
\end{theorem}
\begin{proof}
See Appendix \ref{sec: proofs}.
\end{proof}
Theorem \ref{thm: probability_bound} tells us that a VAE’s $r$-robustness probability can be bounded in terms of: $r$; the Lipschitz constants of the encoder and decoder; the norm of the encoder standard deviation; the dimension of the latent space; and the norm of the input perturbation.
The latter is most important, as it allows us to link the magnitude of input perturbations to the probability of distortions in reconstructions.

The proof leverages the Lipschitz continuity of the decoder network to relate the distances between reconstructed points in $\mathcal{X}$ to the corresponding distances between their latents in $\mathcal{Z}$.
The Lipschitz continuity of the encoder then allows the distribution of distances between samples in latent space --- from perturbed and unperturbed posteriors $q_\phi(\z|\x+\pert)$ and $q_\phi(\z|\x)$ respectively --- to be characterized in terms of distances between VAE inputs.

\begin{figure}[t]
    \centering
    \includegraphics[width=0.25\textwidth, trim = 7mm 7mm 5mm 0, clip=true]{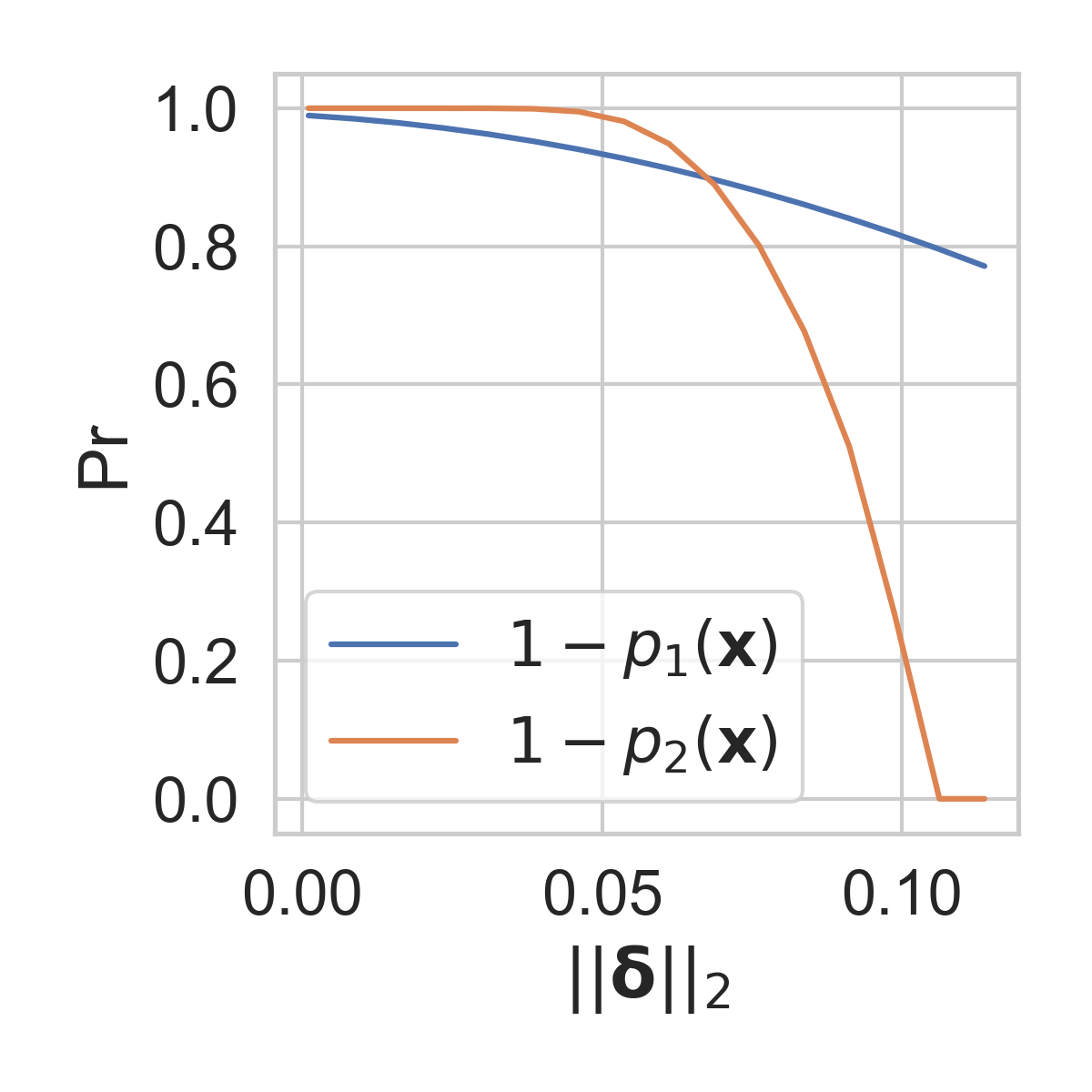}
    \caption[Comparing $p_1(\x)$ and $p_2(\x)$.]{An example of the relative tightness of the bounds in Theorem \ref{thm: probability_bound}, for $a=b=c=r=d_z=5$ and fixed $||\sigma_\phi(\x)||_2=0.1$.
    }
    \label{fig: bound_comparison}
\end{figure}

We note that the distribution of $\ell_2$ distances between these samples is a generalized $\chi^2$ distribution, which has no closed-form CDF \citep{liu2009new}.
The proof therefore employs two tail bounds, Markov's Inequality and a tail bound for standard $\chi^2$ distributions. These varyingly dominate each other in tightness for different $||\pert||_2$ (see Figure \ref{fig: bound_comparison} for a demonstration) and respectively yield $p_1(\x)$ and $p_2(\x)$.

\begin{figure*}[t!]
    \centering
    \subfloat[$M = 5$]{\includegraphics[width=0.24\textwidth]{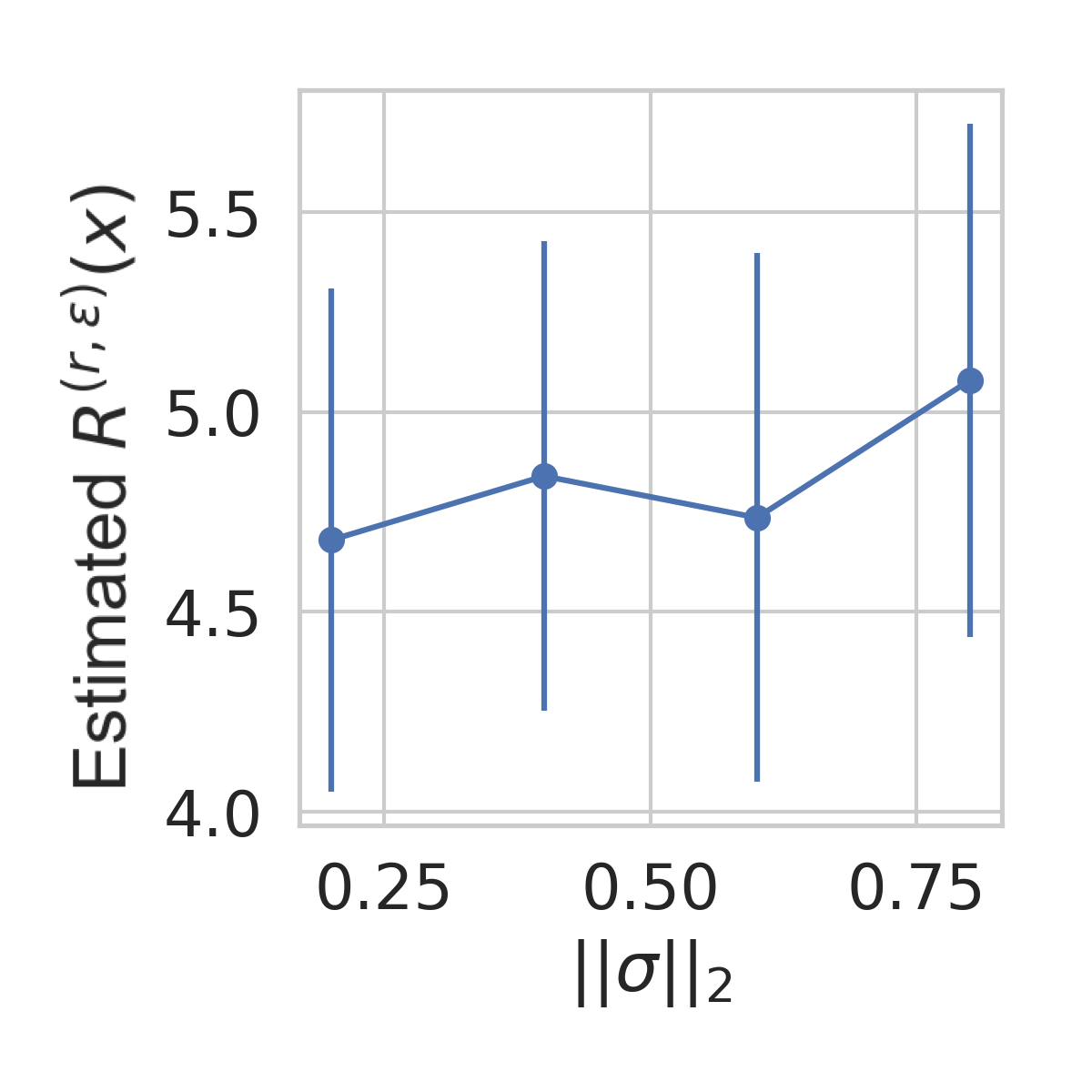}}
    \subfloat[$M = 7$]{\includegraphics[width=0.24\textwidth]{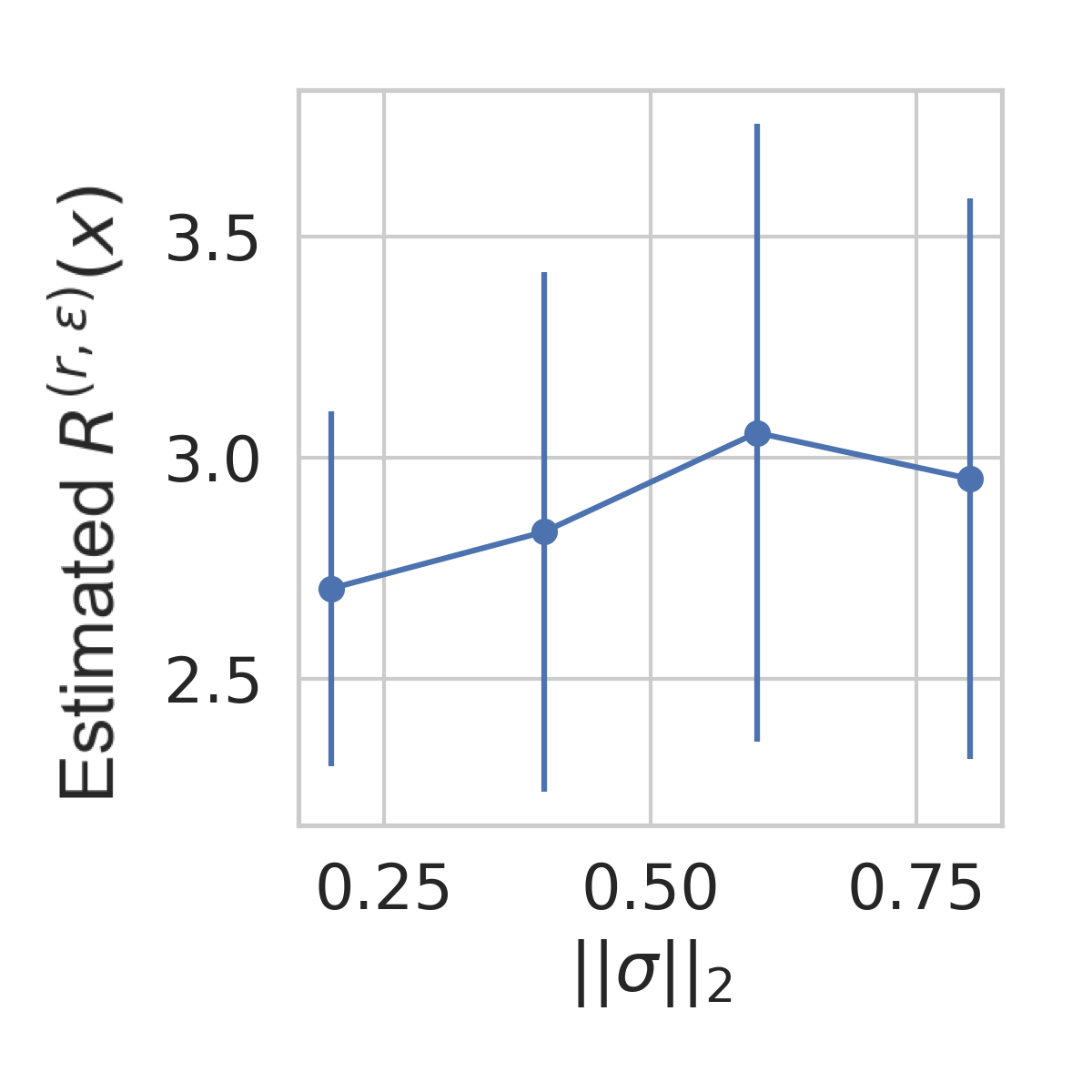}}
    \subfloat[$ M = 10$]{\includegraphics[width=0.24\textwidth]{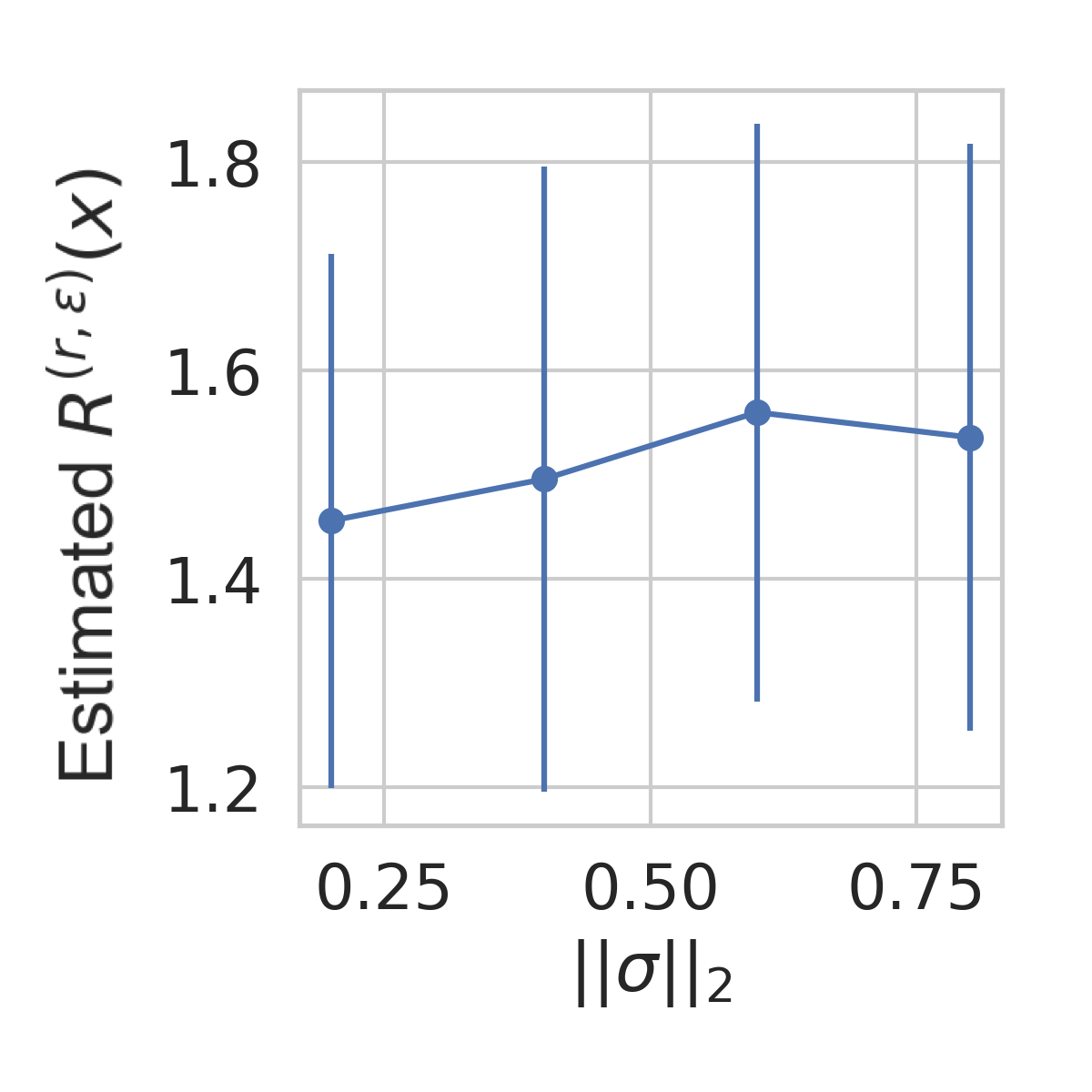}}
    \subfloat[$M = 12$]{\includegraphics[width=0.24\textwidth]{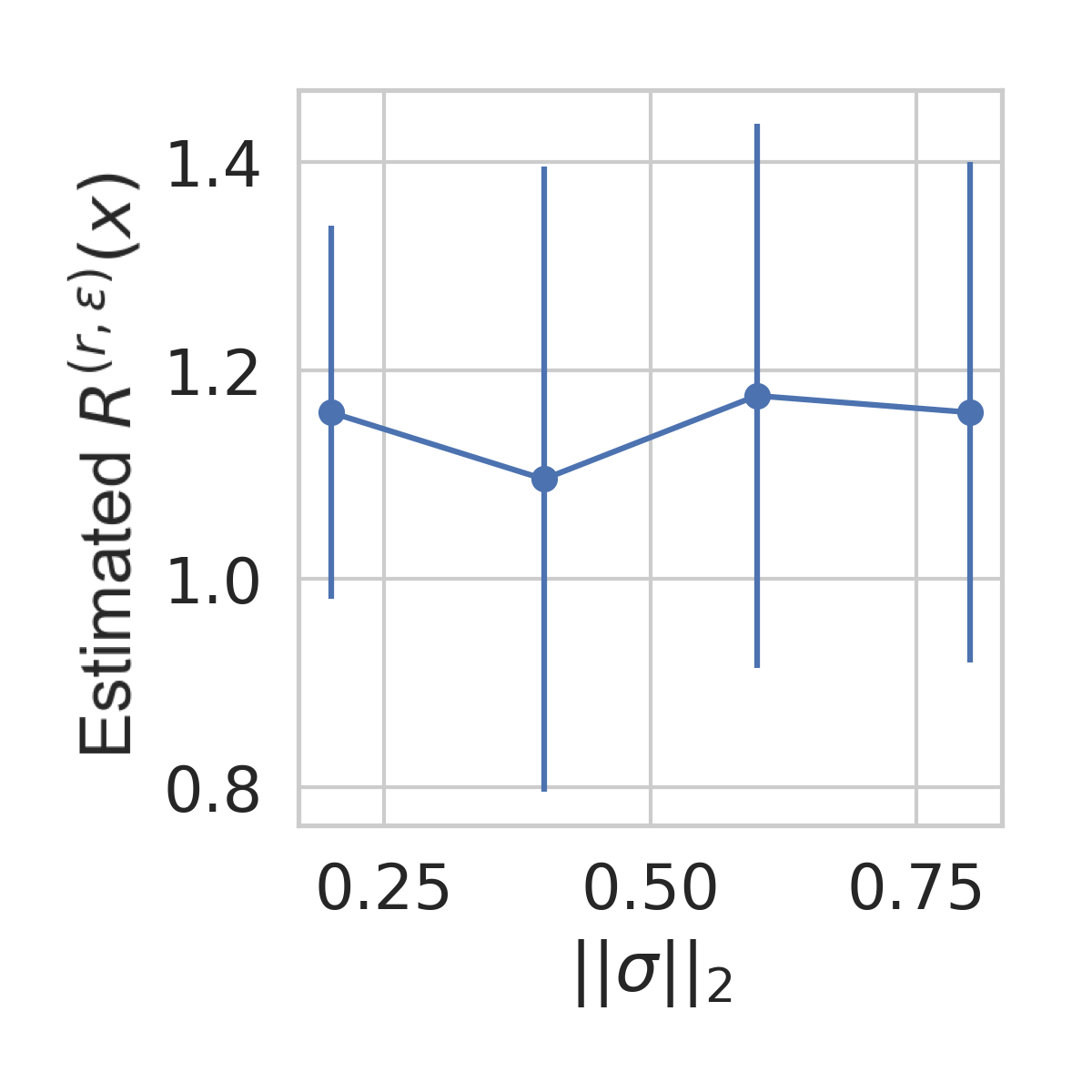}}
    \caption{Estimated $(r, \epsilon)$-robustness margins (see Appendix \ref{sec:app:est_margin}) plotted against the encoder standard deviation norm on MNIST, for fixed $r$ and $\epsilon$ and hyperparameter $||\boldsymbol{\sigma}||_2$ as in Theorem \ref{thm: glob_stoch_margin_bound}. Across Lipschitz constants (denoted by $M$), $||\boldsymbol{\sigma}||_2$ has minimal influence on the estimated robustness margin relative to the choice of Lipschitz constant (compare the ranges across plots). Error bars are the standard deviation over 25 data points.
    }
    \label{fig: R_sigma}
\end{figure*}
\subsection{Bounding The $(r, \epsilon)$-Robustness Margin}
While Theorem \ref{thm: probability_bound} allows for the $r$-robustness probability of a VAE to be lower-bounded for a given input and input perturbation, ideally we would like to guarantee a VAE's robustness at a given input to \emph{all} input perturbations up to some magnitude (for a given $\epsilon$).
The following result provides this guarantee, in terms of a lower bound on the $(r, \epsilon)$-robustness margin.

\begin{lemma}[Margin Bound]
\label{lem: margin_bound}
Given the assumptions of Theorem \ref{thm: probability_bound} and some $\epsilon \in [0, 1)$, the $(r, \epsilon)$-robustness margin of this VAE on input $\x$,
\[R^{(r, \epsilon)}(\x) \geq \max \left\{m_1(\x), m_2(\x) \right\}\]
for
\begin{align*}
    \quad m_1(\x) & := \frac{-4c||\sigma_\phi(\x)||_2 + \sqrt{\Delta}}{2\left(c^2+b^2\right)}, \\
\Delta & := \left(4c||\sigma_\phi(\x)||_2\right)^2 \\
    & \quad -4\left(c^2+b^2\right)\left(4||\sigma_\phi(\x)||_2 - (1-\epsilon) \left(\frac{r}{a}\right)^2\right),
\end{align*}
and $m_2(\x) := \sup \left\{||\pert||_2 : p_2(\pert, \x) \leq (1-\epsilon) \right\},$
where $p_2(\pert, \x)$ is as in Theorem \ref{thm: probability_bound}.\footnote{We augment the listed arguments of $p_2$ to make explicit the dependence on $\pert$.}
\end{lemma}
\begin{proof}
See Appendix \ref{sec: proofs}.
\end{proof}

Lemma \ref{lem: margin_bound} shows that we can lower bound the radius $R^{(r, \epsilon)}$ about $\x$ within which no input perturbation can undermine $(r, \epsilon)$-robustness; when at least one of $m_1(\x)$ and $m_2(\x)$ is positive, robustness can be certified.
The proof exploits the relationship --- established in Theorem \ref{thm: probability_bound} --- between the $r$-robustness probability and the magnitude of input perturbations, finding the largest input perturbation norm such that our lower bound on the $r$-robustness probability still exceeds $\epsilon$.

\subsection{A Global $(r, \epsilon)$-Robustness Margin}

We now extend Lemma~\ref{lem: margin_bound} to provide a \emph{global} margin, which requires bounding $R^{(r, \epsilon)}(\x)$ from below for all $\x \in \mathcal{X}$. This can be done
either by upper-bounding the encoder standard deviation, since the lower bound on the $(r, \epsilon)$-robustness margin from Lemma \ref{lem: margin_bound} is monotonically decreasing in $||\sigma_\phi(\x)||_2$, or by lifting the input dependence entirely, by letting $\sigma_\phi(\x)=\boldsymbol{\sigma} \in \mathbb{R}^{d_z}_{\geq 0}$ be a hyperparameter. Since the derivation when using an upper-bound is equivalent to setting the encoder standard deviation as a hyperparameter, we focus on the latter. Fixing the encoder standard deviation can be done either during training --- since VAEs can be trained with a fixed encoder standard deviation without serious degradation in performance~\citep{Ghosh2020} --- or afterwards, since all that matters to the bound is the value of $\boldsymbol{\sigma}$ at test time.

\begin{theorem}
    [Global Margin Bound]
    \label{thm: glob_stoch_margin_bound}
    Given the assumptions of Lemma \ref{lem: margin_bound}, but with $\sigma_\phi(\x)=\boldsymbol{\sigma} \in \mathbb{R}^{d_z}_{\geq 0}$, the $(r, \epsilon)$-robustness margin of this VAE for all inputs is
    \[R^{(r, \epsilon)} \geq \max \left\{m_1, m_2 \right\}\]
    for
    \[m_1 := \frac{\sqrt{-\left(4||\boldsymbol{\sigma}||_2^2-(1-\epsilon)\left(\frac{r}{a}\right)^2 \right)}}{b}\]
    and $m_2 := \sup \left\{||\pert||_2 : p_2(\pert) \leq (1-\epsilon)\right\}$,
    where $p_2$ is as in Theorem \ref{thm: probability_bound}, but $u := \frac{\left(\frac{r}{a}-b||\pert||_2\right)^2}{4||\boldsymbol{\sigma}||_2^2}.$
    \end{theorem}
\begin{proof}
See Appendix \ref{sec: proofs}.
\end{proof}

This result provides guarantees solely in terms of parameters we can choose \emph{ahead} of training, namely the Lipschitz constants of the networks and $\boldsymbol{\sigma}$, the fixed value of the encoder standard deviation. This importantly distinguishes ours from previous work, which has only provided robustness bounds based on intractable model characteristics that must be empirically estimated \emph{after training}~\citep{alex2020theoretical}.

Note that to further investigate the impact of using fixed encoder standard deviations, we train VAEs with the encoder standard deviation set as a hyperparameter.  As shown in Figure~\ref{fig: R_sigma}, we find the Lipschitz constants of the encoder and decoder networks to be most determinative for robustness, with the value of $||\boldsymbol{\sigma}||_2$ being of lesser importance.

\section{IMPLEMENTATION}

In the last section, we introduced
guarantees on robustness assuming the Lipschitz constants of a VAE's networks.
We now consider how to train a VAE in a manner that ensures these guarantees are met.

Letting $\mathcal{F}$ be the set of functions that can be learned by an unrestricted neural network, and $\mathcal{L}_M \subset \mathcal{F}$ be the (further restricted) subset of $M$-Lipschitz continuous functions associated with the sets of neural network parameters $\mathcal{L}^\theta_M,\mathcal{L}^\mathcal{\phi}_M$,
our constraint can be thought of simply as replacing the standard VAE objective in \eqref{eq: likelihood_lower_bound}
with the modified objective
\[\argmax_{\theta, \phi \in \mathcal{L}^\theta_M,\mathcal{L}^\mathcal{\phi}_M} \quad
    \mathbb{E}_{q_\phi(\z|\x)}[\log p_\theta(\x|\z)] - \kl{q_\phi(\z|\x)}{p(\z)}.\]
Referring to VAEs trained this way as \emph{Lipschitz-VAEs}, the question becomes how to enforce this objective.
Using \cite{anil2018sorting}, we focus on fully-connected networks, although similar ideas extend to other architectures \citep{li2019preventing}. First, note that if layer $l$ has Lipschitz constant $M_l$, then the Lipschitz constant of the entire network is $M=\prod_{l=1}^L M_l$ \citep{szegedy2013intriguing}.
For an $L$-layer fully-connected neural network to be $M$-Lipschitz, it thus suffices to ensure that each layer has Lipschitz constant $M^{\frac{1}{L}}$.
If we choose the network non-linearity $\varphi_l(\cdot)$ to be $1$-Lipschitz, and ensure that linear transformation $\mathbf{W}_l$ is also $1$-Lipschitz,
then a Lipschitz constant of $M^{\frac{1}{L}}$ in each layer follows from scaling the outputs of each layer by $M^{\frac{1}{L}}$.

\begin{algorithm}[t]
\SetKwInOut{Input}{Input}
\SetKwInput{Requires}{Requires}
\SetKwFunction{BjorckFunc}{Bj\"{o}rckOrthonormalize}
\SetKwBlock{Forward}{Forward pass}{Forward pass}
\SetKwBlock{BjorckBlock}{Bj\"{o}rckOrthonormalize}{Bj\"{o}rckOrthonormalize}
\BjorckBlock{
\For{$k=1, \ldots, K$}{
$\mathbf{W}_l^{(k+1)} \leftarrow$ \\
$\mathbf{W}_l^{(k)} \left(I + \frac{1}{2}Q^{(k)} + \ldots + (-1)^p \begin{pmatrix} -0.5 \\
p\end{pmatrix} (Q^{(k)})^p \right)$\\
where $Q^{(k)}:=I - \left(\mathbf{W}_l^{(k)}\right)^\intercal \mathbf{W}_l^{(k)}$, and $K$ and $p$ are hyperparameters.}
}
\Input{Data point $\x$}
\KwResult{Network output $\mathbf{h}_L$}
\Requires{Lipschitz constant $M$}
\nonl
\Forward{
\nonl $\mathbf{h}_0 \leftarrow \x$ \;

\nonl
\For{$l=1, \ldots, L$}{
\nonl $\mathbf{W}_l \leftarrow$ \BjorckFunc{$\mathbf{W}_l$}\; \\
pre-activation $\leftarrow$ $M^{\frac{1}{L}} \mathbf{W}_l\mathbf{h}_{l-1}$ \; \\
$\mathbf{h}_l \leftarrow \mathrm{GroupSort}(\text{pre-activation})$\;
}}
\caption[A Lipschitz continuous neural network's forward pass during training.]{The forward pass in a Lipschitz-VAE's encoder or decoder network.}
\label{alg: training_forward_pass}
\end{algorithm}

Building on this, our approach to controlling the Lipschitz continuity of VAE encoders and decoders can be seen in Algorithm \ref{alg: training_forward_pass}.
The key components are Bj\"{o}rck Orthonormalization, which ensures each layer's linear transformation is $1$-Lipschitz, and the $\mathrm{GroupSort}$ non-linearity from \cite{anil2018sorting}, which is also $1$-Lipschitz. See Appendix \ref{app:lipschitz_implement} for more details.

\section{RELATED WORK}

\paragraph{Certifiable Robustification} Prior work on robustifying models to adversarial attacks can be delineated into a) techniques providing robustness to known types of attack empirically, and b) certifiable techniques providing provable robustness under assumptions.
\citet{cohen2019certified} argues certifiable techniques should be favored since empirical findings of robustness are predicated on a choice of attack and thus cannot indicate effectiveness against other known or as yet unknown attacks.
Indeed, we previously noted instances where empirically-led techniques seemed to induce robustness but were subsequently undone by later-developed attacks \citep{athalye2018obfuscated, uesato2018adversarial}.

\paragraph{Certifiable Robustness In Classifiers} Given their advantages, certifiable robustification techniques have already been developed for classifiers, where approaches employing Lipschitz continuity are particularly illustrative.
In particular, \cite{hein2017formal, tsuzuku2018lipschitzmargin, anil2018sorting,  yang2020} use Lipschitz continuity to provide certified robustness margins for classifiers.
We note, however, that in that setting one does not need to handle the probabilistic aspects and continuous changes that one finds in VAEs.

\paragraph{Robustness In VAEs} \cite{willetts2021improving} argues that the susceptibility of a VAE to adversarial perturbations depends on how much the encoder $q_\phi(\z|\x)$ can be changed through changes in input $\x$, and how much reconstruction $g_\theta(\z)$ can be changed through changes in the latent variable $\z$. \cite{cemgil2020autoencoding} similarly holds that adversarial examples are possible in VAEs due to ``non-smoothness'' in the encoding-decoding process, relating this to dissimilarity between a VAE's reconstructions of its reconstructions.
\cite{willetts2021improving} targets greater smoothness heuristically, controlling the noisiness of the VAE encoding process so that ``nearby'' inputs correspond to ``nearby'' latent variables and changes in $q_\phi(\z|\cdot)$ induced by input perturbations have little effect on reconstructions $g_\theta(\z)$.
Separately, \cite{alex2020theoretical} proposes $(r, \epsilon)$-robustness and obtains an approximate bound on the $(r, \epsilon)$-robustness margin, allowing the robustness of VAEs to be assessed.
That work assumes, however, that input perturbations only affect the encoder mean and not its standard deviation.
That work also only allows for the assessment of the robustness of \textit{already trained} VAEs, and unlike our methods does not directly enforce \textit{guaranteed} robustness.

\begin{figure*}[t!]
    \centering
    \subfloat[Lemma \ref{lem: margin_bound}: MNIST]{\includegraphics[width=0.28\textwidth]{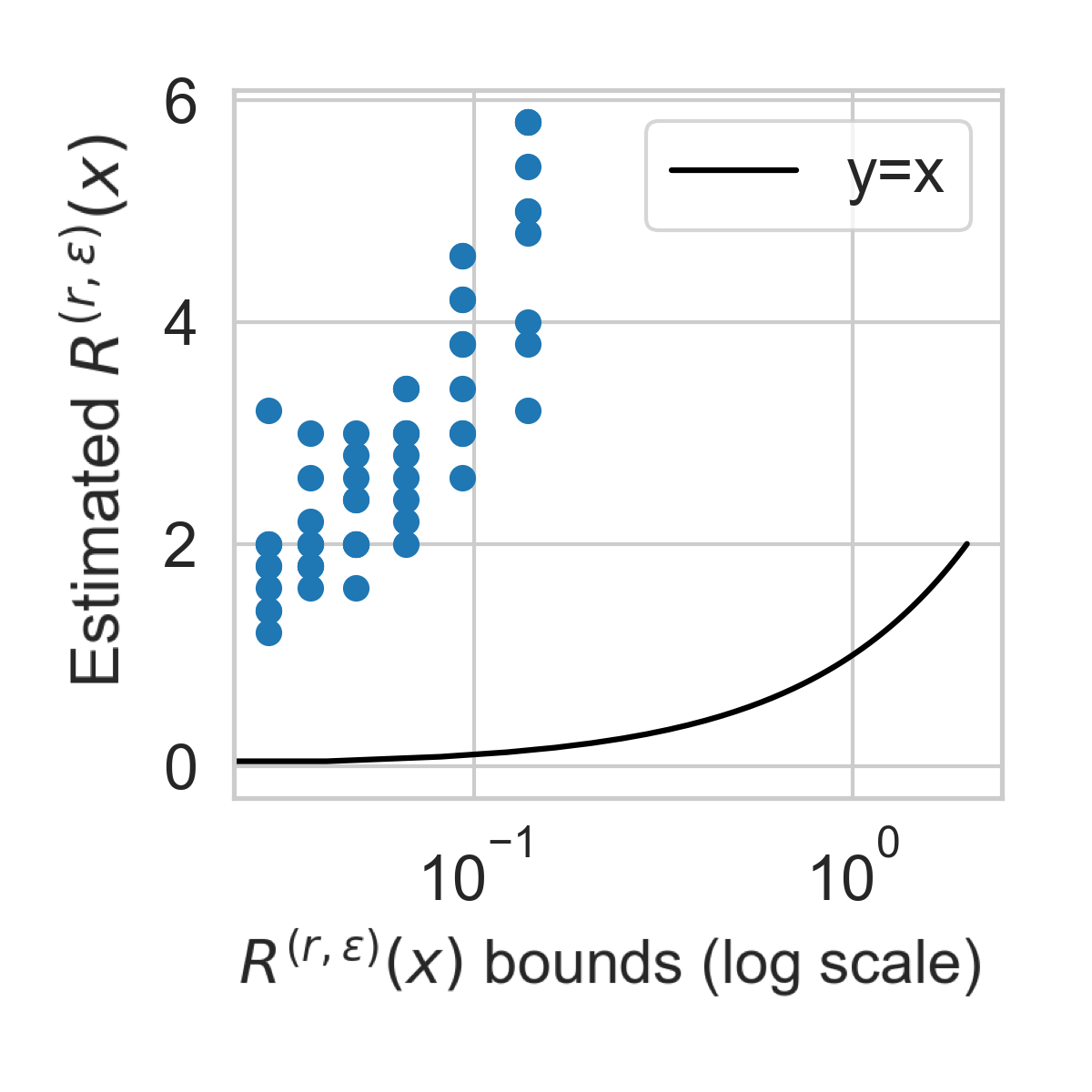}}
     \subfloat[Lemma \ref{lem: margin_bound}: Fashion-MNIST]{\includegraphics[width=0.28\textwidth]{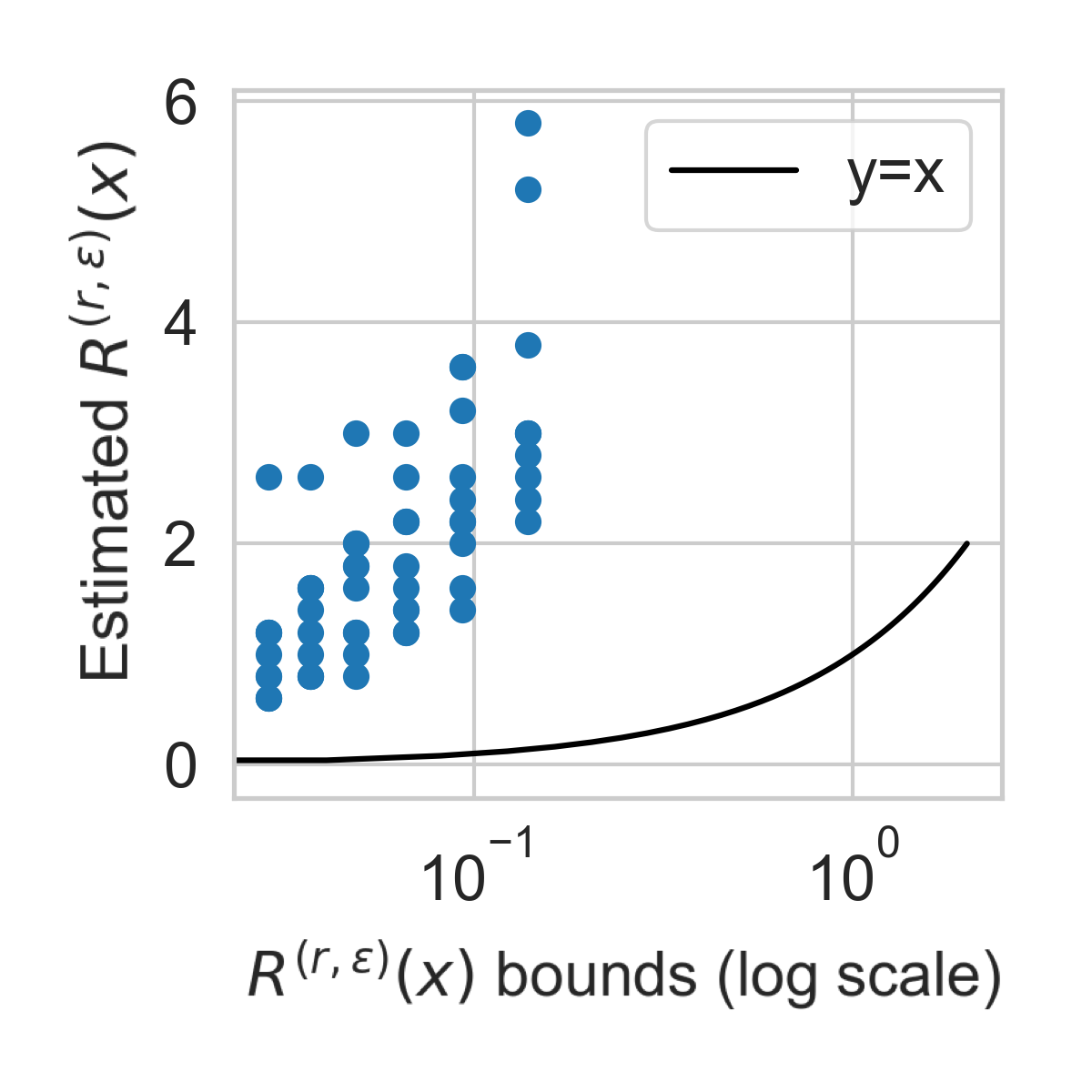}}
     \subfloat[Theorem \ref{thm: glob_stoch_margin_bound}: MNIST]{\includegraphics[width=0.28\textwidth]{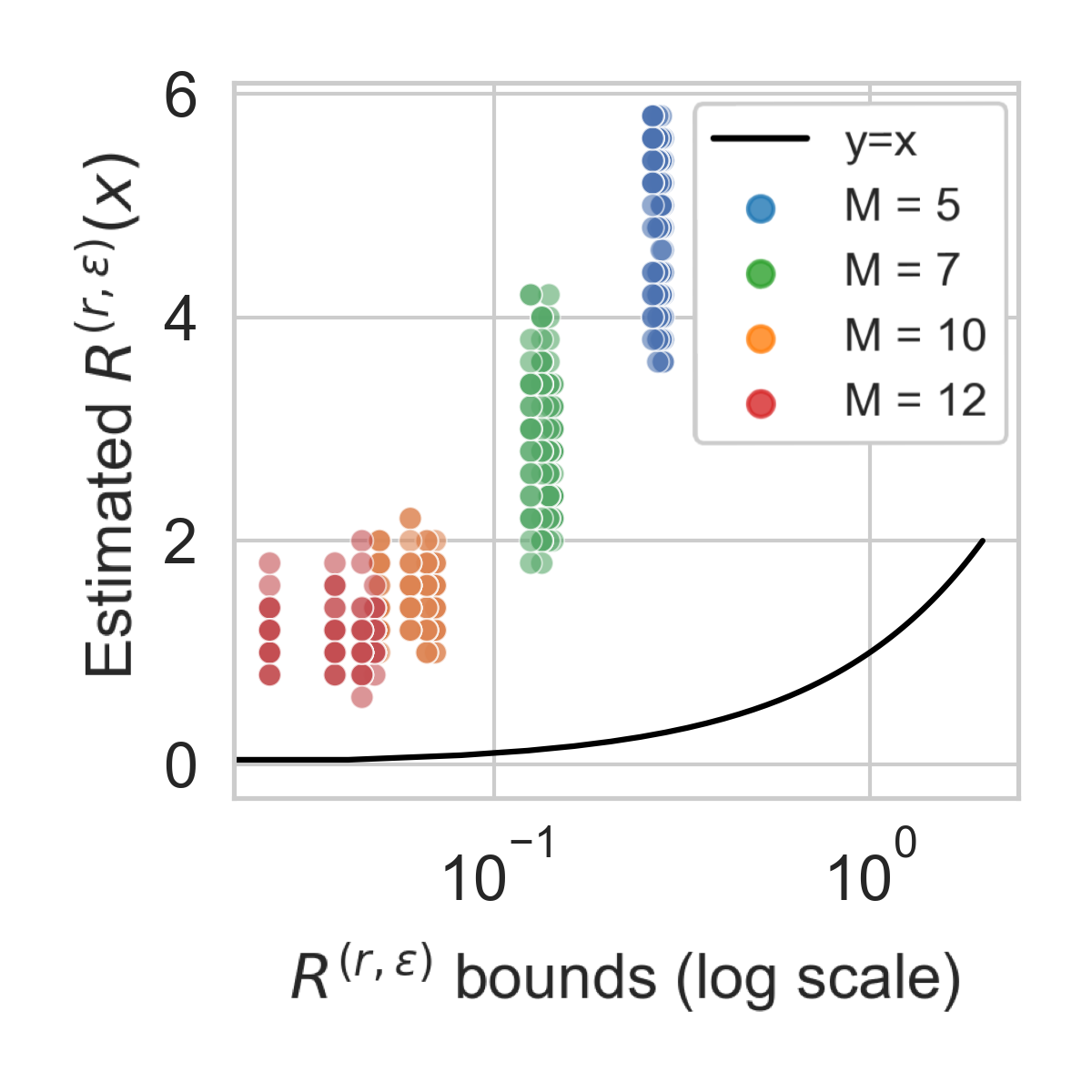}}
    \caption[Comparing $(r, \epsilon)$-robustness margin estimates.]{Estimated $(r, \epsilon)$-robustness margins plotted against the lower bounds on these margins from Lemma \ref{lem: margin_bound}, for networks trained on \textbf{[Left]} MNIST and \textbf{[Center]} Fashion-MNIST.  \textbf{[Right]} The same plot for the bound in Theorem \ref{thm: glob_stoch_margin_bound} on MNIST, for fixed $||\boldsymbol{\sigma}||_2 \in \{0.06, 0.13,0.19,0.25\}$ and Lipschitz constants $M \in \{5, 7, 10, 12\}$. We plot $y=x$ to illustrate the correctness of the bounds, and use $r=8$ and $\epsilon=0.5$ throughout.
    }
    \label{fig: empirical_theoretical_comparison}
\end{figure*}

\section{EXPERIMENTS}
\label{sec: experiments}

Our aim now is to establish that our theoretical results allow certifying and guaranteeing VAE robustness in practice. We would also like to verify that Lipschitz continuity constraints can endow VAEs with greater robustness to adversarial inputs than standard VAEs.

\paragraph{Experimental Setup}
We pick a latent space with dimension $d_z=10$ (unless otherwise stated) and use the same architecture across experiments: encoder mean $\mu_\phi(\cdot)$, encoder standard deviation $\sigma_\phi(\cdot)$ and deterministic component of the decoder $g_\theta(\cdot)$ are all three-layer fully-connected networks with hidden dimensions $512$ (for more details, see Appendix~\ref{app: net_arch}). Following \cite{anil2018sorting}, we start with $K=3$ Bj\"{o}rck Orthonormalization iterations before setting $K=50$ to finetune to convergence (using $p=1$ throughout).

\begin{figure*}[t!]
    \centering
    \includegraphics[width=0.49\textwidth]{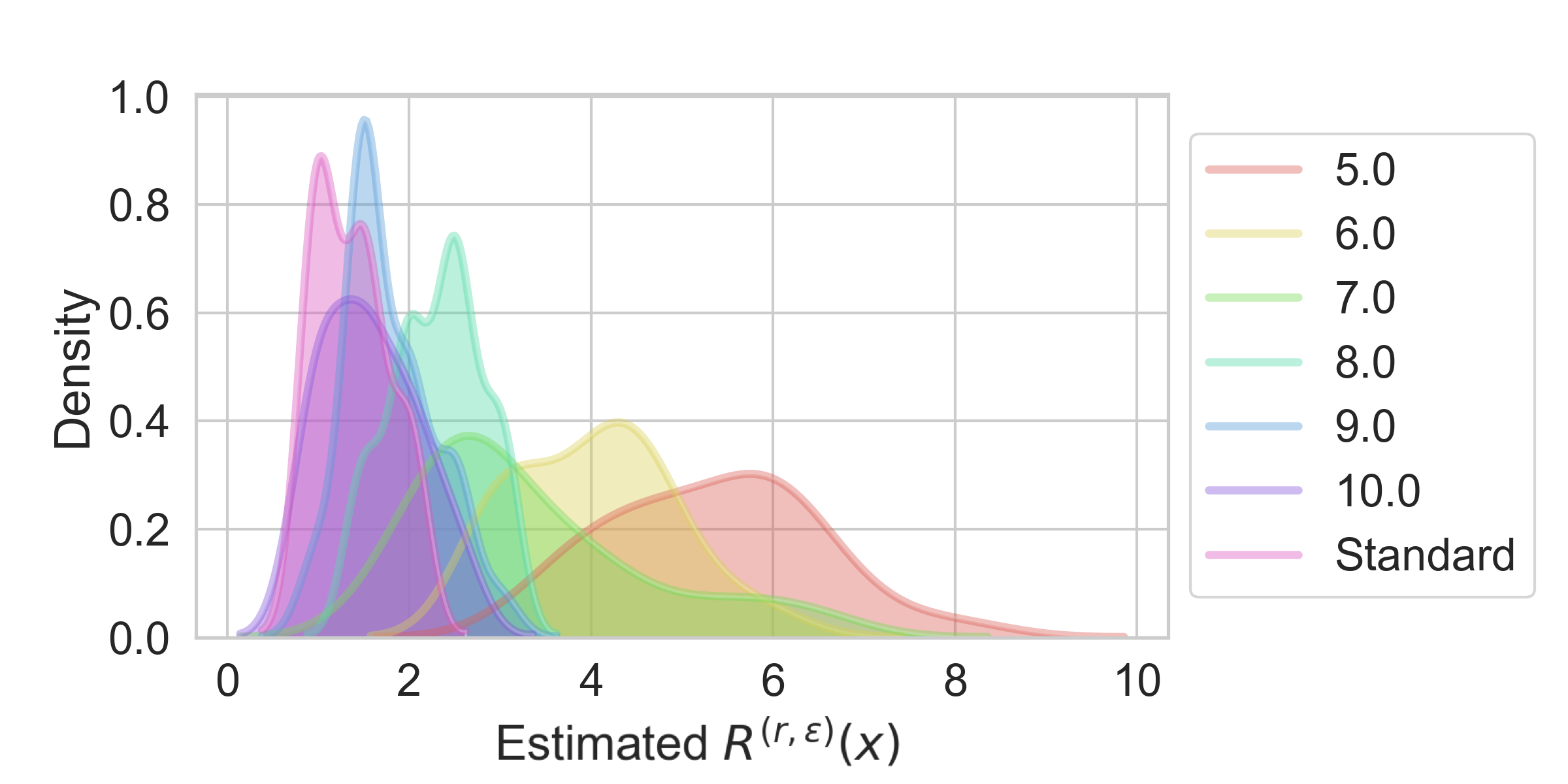}
    \includegraphics[width=0.31\textwidth]{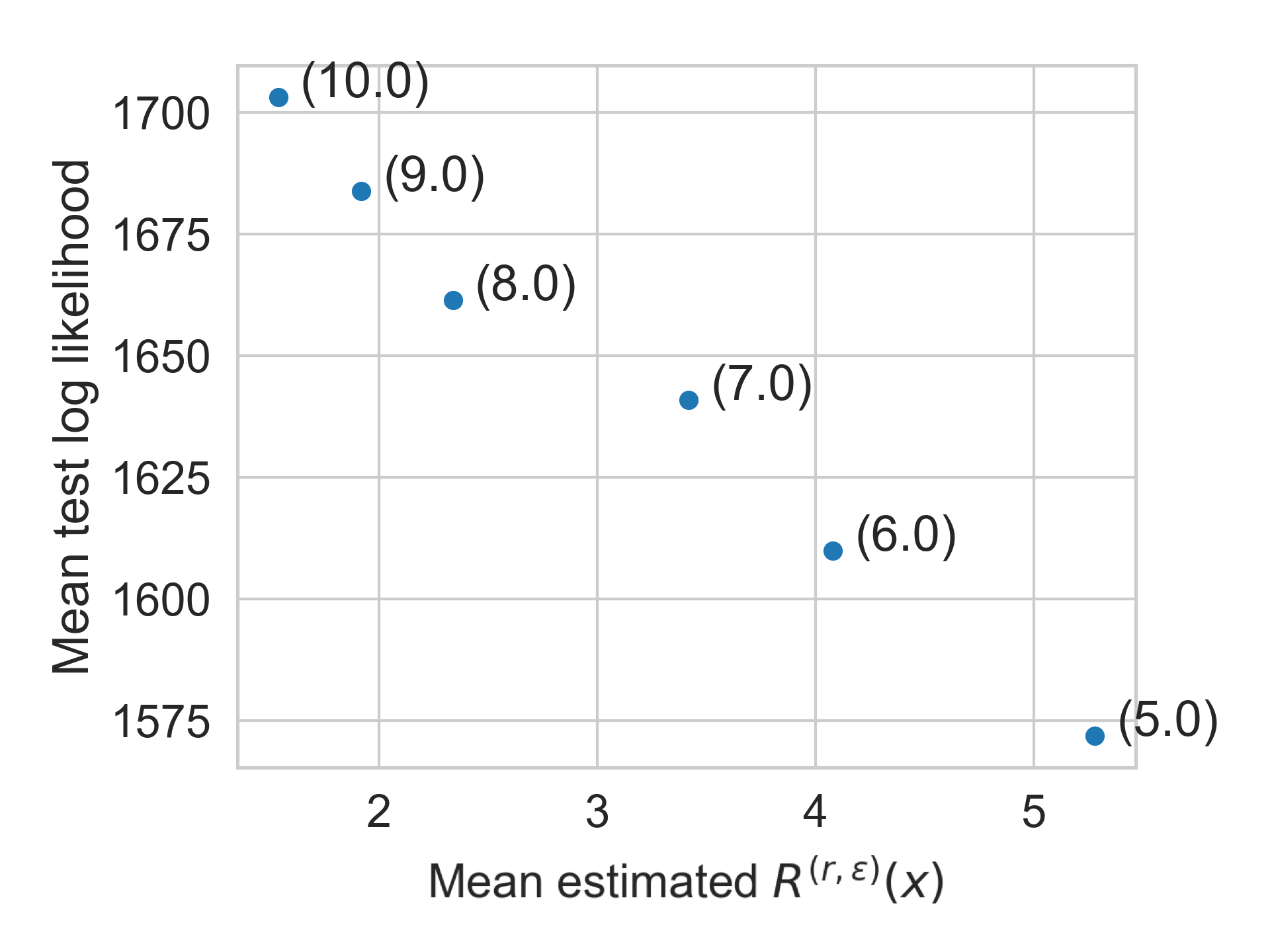}
    \caption{\textbf{[Left]} $(r, \epsilon)$-robustness margins $R^{(r, \epsilon)}(\x)$ estimated using maximum damage attacks on a randomly-selected collection of MNIST data points in Lipschitz- and standard VAEs, for $r=8$, $\epsilon=0.5$, and $||\boldsymbol{\sigma}||_2=0.1$. For all Lipschitz constants considered, Lipschitz-VAEs exhibit larger $(r, \epsilon)$-robustness margins on average than a standard VAE, demonstrating the empirical robustness of Lipschitz-VAEs. Larger $(r, \epsilon)$-robustness margins also correlate with smaller Lipschitz constants, as predicted by our bounds. \textbf{[Right]} The empirical relationship between a Lipschitz-VAE's reconstruction performance, measured by the mean log likelihood of reconstructions on the MNIST test set, and its mean robustness margin, by Lipschitz constant (in parentheses).
    }
    \label{fig: empirical_robustness_margins}
\end{figure*}

\paragraph{Validating Certifiable Robustness}
As a sanity check, we first empirically validate that our bounds in Lemma \ref{lem: margin_bound} and Theorem \ref{thm: glob_stoch_margin_bound} allow us to provide the advertised absolute robustness guarantees. Namely, for a given $r$, $\epsilon$, and Lipschitz-VAE, we compute $\max\{m_1(\x), m_2(\x)\}$ and $\max\{m_1, m_2\}$, for Lemma \ref{lem: margin_bound} and Theorem \ref{thm: glob_stoch_margin_bound} respectively, on a randomly-selected sample from MNIST and Fashion-MNIST (see Figure~\ref{fig: empirical_theoretical_comparison}).

This experiment empirically validates our bounds, since in all instances the estimated $(r, \epsilon)$-robustness margins (see the following section for estimation) are larger than our corresponding theoretical bounds on these margins.
We also see that the bounds on the $(r, \epsilon)$-robustness margin are strictly positive, providing a priori guarantees of robustness when choosing a fixed (or bounded) encoder standard deviation and encoder and decoder Lipschitz constants as in Theorem \ref{thm: glob_stoch_margin_bound}.
Our results demonstrate the existence of Lipschitz-VAEs for which meaningful robustness can be certified, a priori.

Though Figure~\ref{fig: empirical_theoretical_comparison} suggests our bounds may often be relatively loose, this is very much consistent with applications of Lipschitz continuity constraints in other settings~\citep{cohen2019certified}.
This looseness is perhaps unavoidable, since an a priori theoretical guarantee of robustness is an extremely strong requirement. %
As such, our approach is useful in scenarios where robustness must be absolutely guaranteed, even if at times the level of robustness that is guaranteed is lower than the level observed in practice.

\begin{figure*}[t!]
\captionsetup[subfloat]{farskip=-1pt,captionskip=-1pt}
\centering
        \subfloat[Standard VAE]{\includegraphics[width=0.3\textwidth]{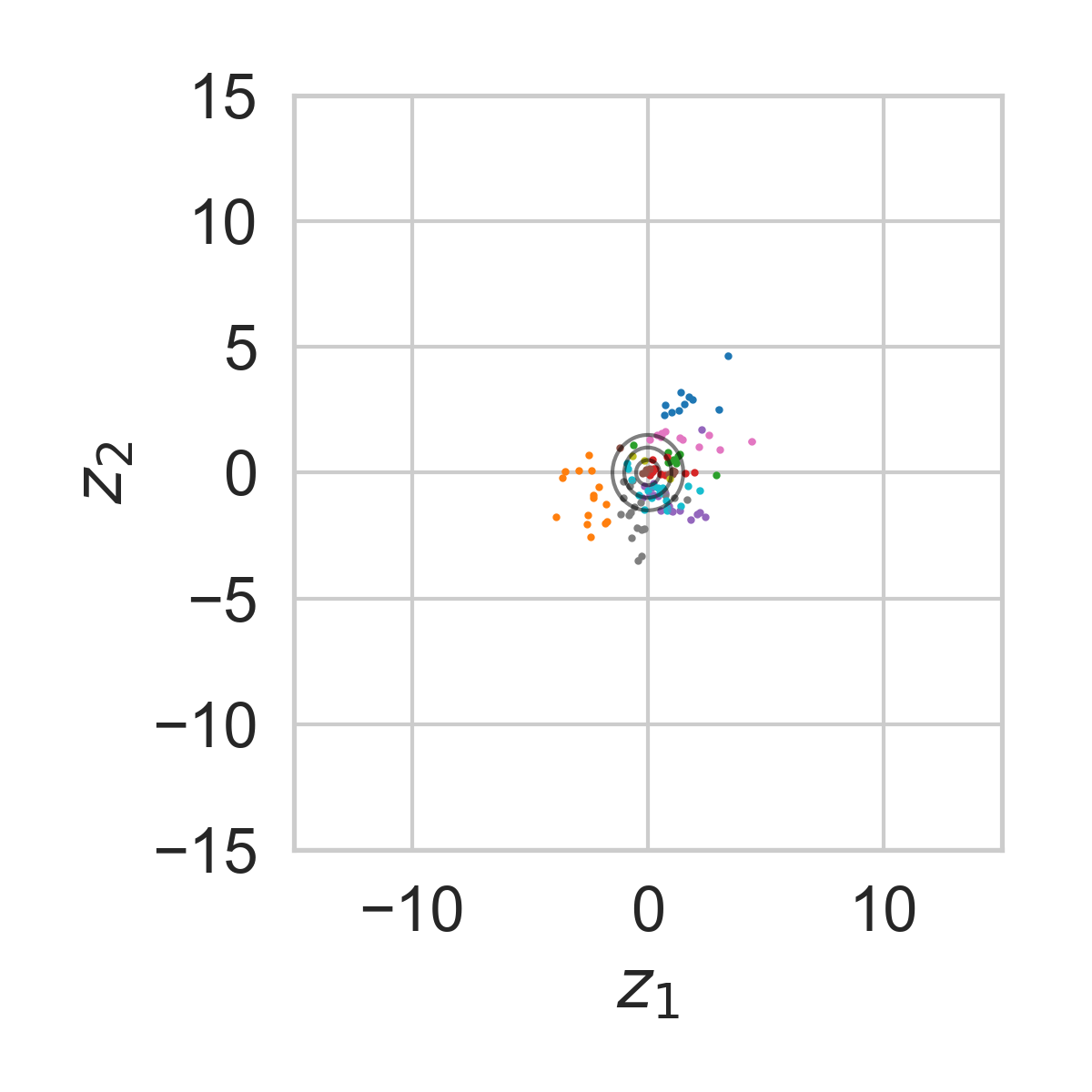}}
        \subfloat[Lipschitz-VAE, $M = 10$]{\includegraphics[width=0.3\textwidth]{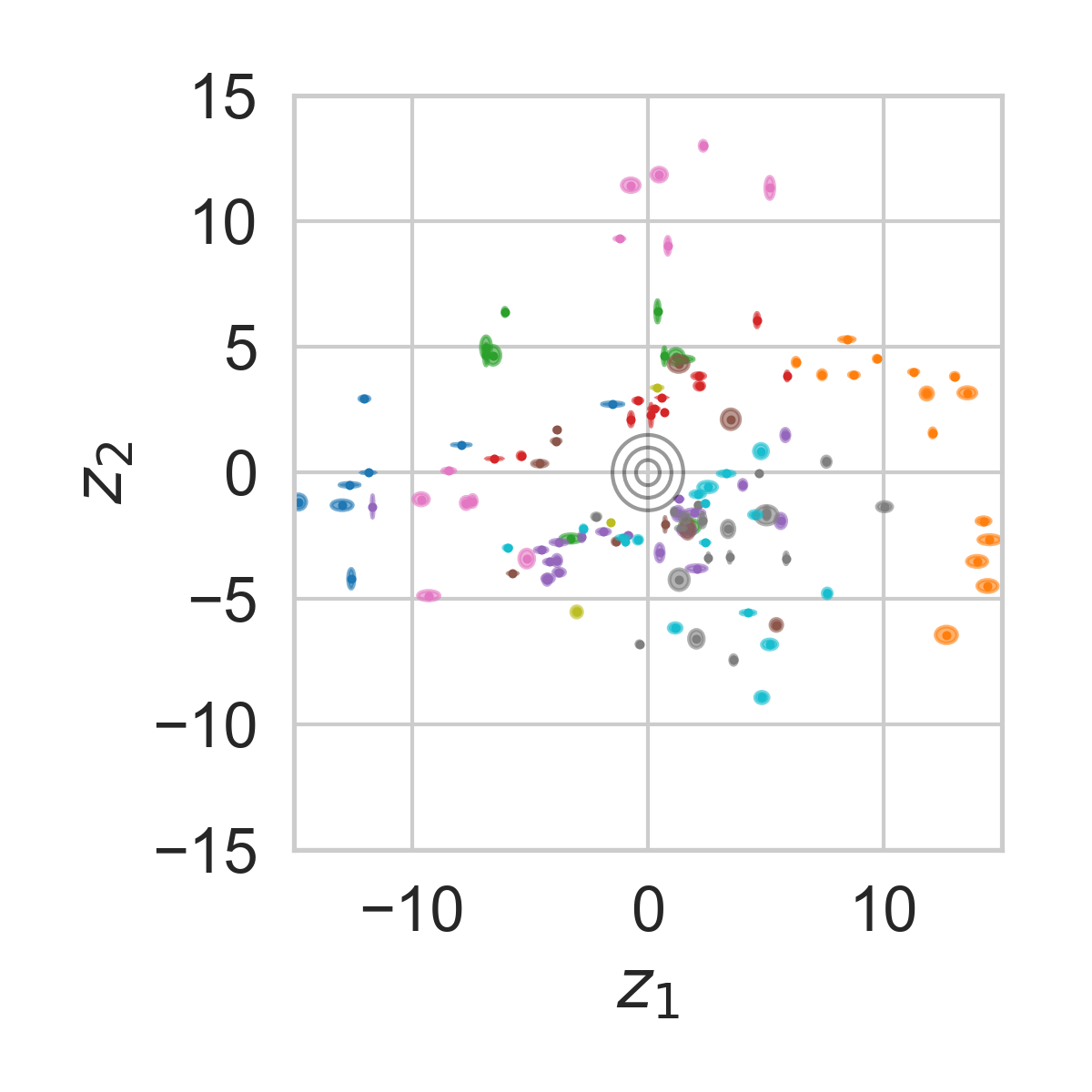}}
        \subfloat[Lipschitz-$\beta$-VAE, $M = 10$, $\beta$=$5$]{\includegraphics[width=0.3\textwidth]{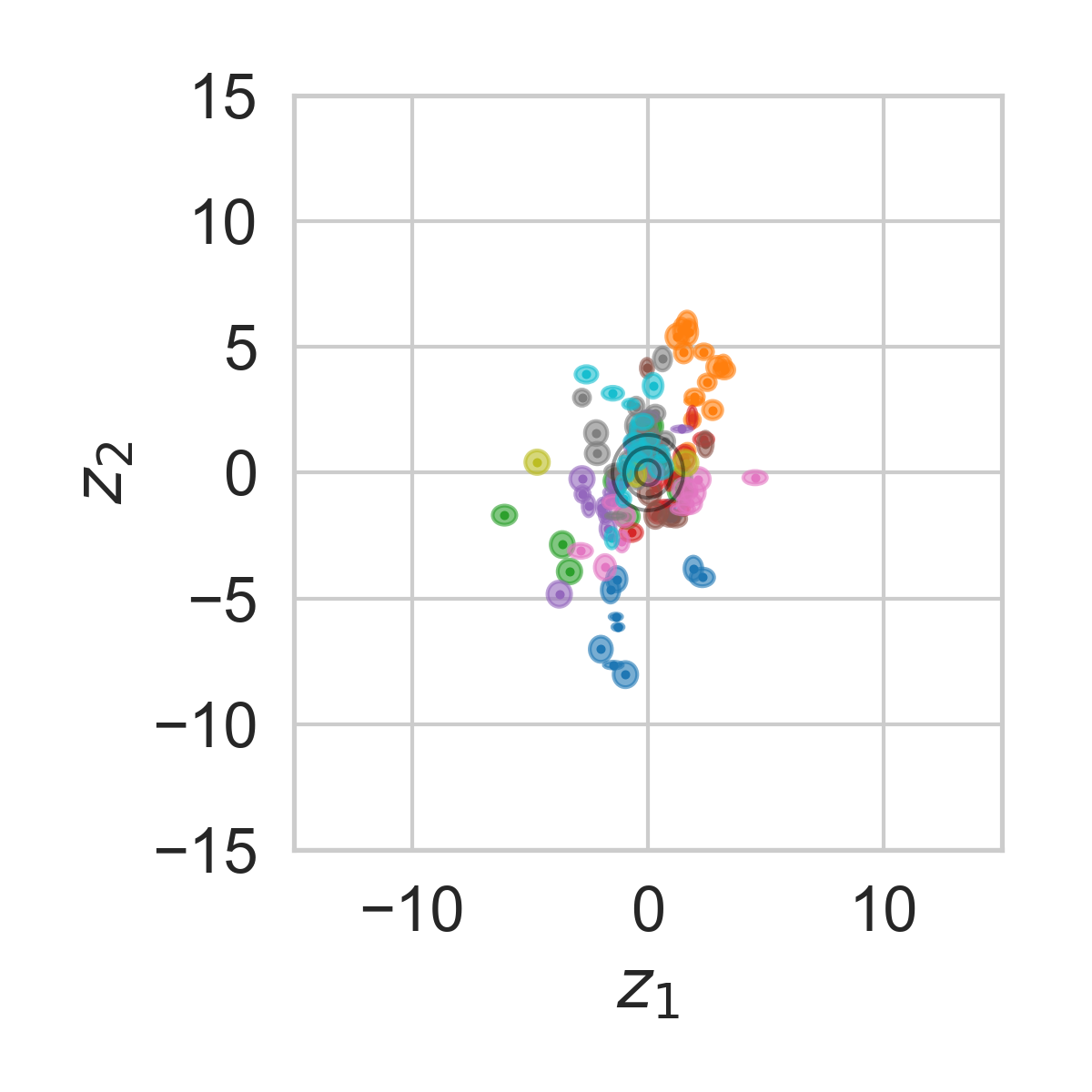}}
        \caption[The encoders learned by different types of VAE.]{Learned encodings for different types of VAE on MNIST. A colored ellipse represents the posterior $q_\phi(\z|\x_i)$ for a single $\x_i$. The prior, $p(\z)=\mathcal{N}({\z};{\mathbf{0}},{\mathbf{I}})$, is overlaid in black for one, two and three standard deviations. Lipschitz-VAEs have encoders that are dispersed in latent space, in contrast with the learned encoder of a standard VAE. Upweighting the KL term in \eqref{eq: likelihood_lower_bound}, as in a $\beta$-VAE \citep{Higgins2017betaVAELB}, changes this behaviour.}
    \label{fig: learned_latent_spaces}
\end{figure*}

\paragraph{Comparing Empirical Robustness}
We next empirically assess the $(r, \epsilon)$-robustness margins of Lipschitz-VAEs using the approach of~\cite{alex2020theoretical}, leveraging maximum damage attacks (see Appendix~\ref{sec:app:est_margin}).
Assuming no defects in the optimization of \eqref{eq: max_damage_attack} and access to infinite samples from the encoder, if a maximum damage attack cannot identify a $\pert^* \leq c$ such that $\prob{||g_\theta(\z_{\pert^*})-g_\theta(\z_{\neg \pert^*})||_2 \leq r} \leq \epsilon$,
then we can rest assured that the $(r, \epsilon)$-robustness margin of the VAE on input $\x$ is at least $c$. %
If (for fixed $r$ and $\epsilon$) one VAE's estimated $(r, \epsilon)$-robustness margins are consistently larger than another's,
this strongly suggests that the former is more robust.

In Figure \ref{fig: empirical_robustness_margins} [left], we estimate the $(r, \epsilon)$-robustness margins of several Lipschitz- and standard VAEs on a randomly-selected collection of data points from MNIST. On the same inputs, and for all Lipschitz constants considered, Lipschitz-VAEs exhibit larger estimated $(r, \epsilon)$-robustness margins on average.
Figure \ref{fig: empirical_robustness_margins} [left] also validates an implication of our theory, namely that a VAE's $(r, \epsilon)$-robustness margins should monotonically increase as we decrease its Lipschitz constants.

We thus demonstrate that we can manipulate the robustness levels of Lipschitz-VAEs through their Lipschitz constants, fulfilling our objective to develop a VAE whose robustness levels can be controlled \textit{a priori}. %
Note that, using a unit Gaussian prior, we find the useful range of Lipschitz constants for all networks considered to be between about five and ten:
less than this reconstructive performance is excessively impacted, while greater than this Lipschitz-VAEs exhibit robustness comparable to standard VAEs.

\paragraph{Choosing Lipschitz Constants}
Previously, we saw that the $(r, \epsilon)$-robustness margins of a Lipschitz-VAE could be manipulated through its Lipschitz constants, with smaller Lipschitz constants consistently affording greater robustness. In practice, however, robustness might only be one consideration, alongside reconstruction performance, in choosing between VAEs.

To explore these considerations, we plot reconstruction performance against estimated robustness in Figure \ref{fig: empirical_robustness_margins} [right], measuring reconstruction performance as the mean log likelihood achieved, and estimating robustness in terms of $R^{(r, \epsilon)}(\x)$.
Recalling that larger log likelihoods imply better reconstructions, we see that reconstruction performance is negatively correlated with estimated robustness, with behavior on each dimension determined by the Lipschitz constants.

\begin{figure}
    \centering
    \includegraphics[width=0.33\textwidth, trim = 7mm 7mm 5mm 0, clip=true]{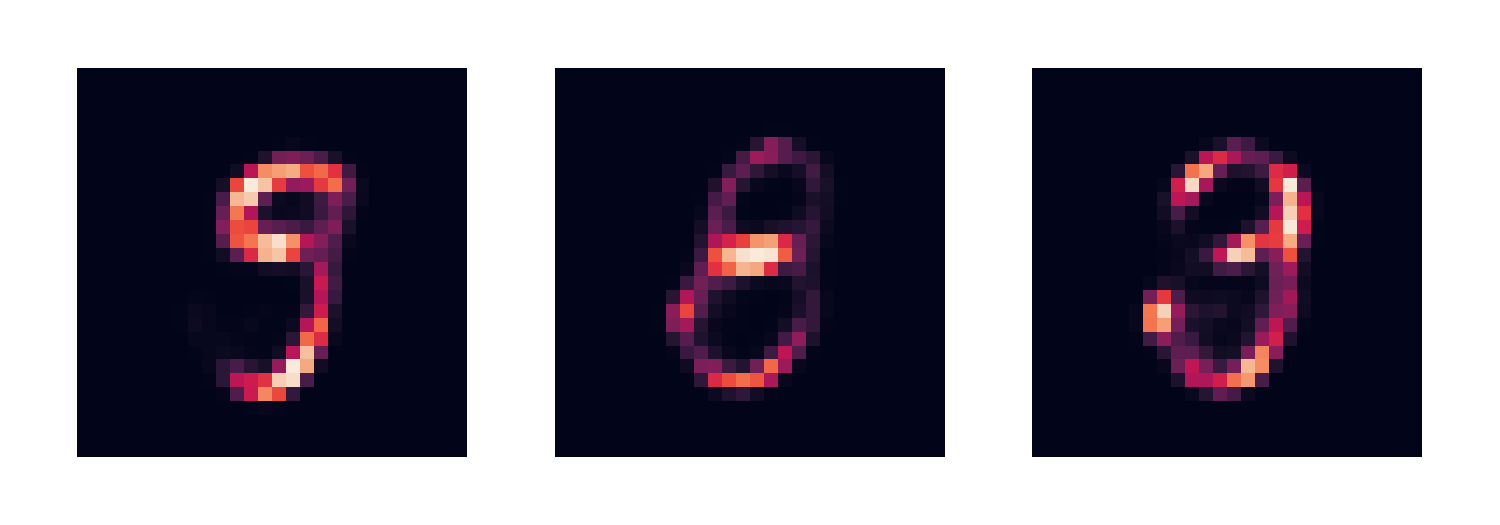}
    \caption[Sample generations.]{Sample generations from a $10$-Lipschitz VAE decoder using noise from a unit Gaussian prior.}
    \label{fig: sample_generations}
\end{figure}

\paragraph{Investigating Learned Latent Spaces}
We now empirically examine whether Lipschitz-VAEs are qualitatively different from standard VAEs in aspects other than robustness, in particular in the latent spaces they learn.
As shown in Figure \ref{fig: learned_latent_spaces}, the aggregate posteriors learned by Lipschitz- and standard VAEs differ in their scale.
The aggregate posterior of a standard VAE is tightly clustered about the prior $p(\z)=\gauss{\z}{\mathbf{0}}{\mathbf{I}}$, but that of a Lipschitz-VAE disperses mass more widely over the latent space.

Though this could be an issue when generating samples from the prior, as the prior and aggregate posterior have little overlap, the remedy to this issue is quite simple.
We find that upweighting the KL term in the VAE objective by hyperparameter $\beta$, as in a $\beta$-VAE \citep{Higgins2017betaVAELB}, mitigates this scaling of the latent space (see Figure \ref{fig: learned_latent_spaces}c).
For details, see Appendix \ref{app:latent_space_explore}; sample generations can also be seen in Figure \ref{fig: sample_generations}.

\section{CONCLUSION}
We have introduced an approach to training VAEs that allows their robustness to adversarial attacks to be guaranteed \textit{a priori}. Specifically, we derived provable bounds on the degree of robustness of a VAE under input perturbation, with these bounds depending on parameters such as the Lipschitz constants of its encoder and decoder networks. We then showed how these parameters can be controlled, enabling our bounds to be invoked in practice and providing an actionable way of ensuring the robustness of a VAE ahead of training.

\bibliography{citations}

\begin{thebibliography}{}

\bibitem[Anil et~al., 2019]{anil2018sorting}
Anil, C., Lucas, J., and Grosse, R. (2019).
\newblock {Sorting Out Lipschitz Function Approximation}.
\newblock In {\em International Conference on Machine Learning}, pages
  291--301.

\bibitem[Athalye et~al., 2018]{athalye2018obfuscated}
Athalye, A., Carlini, N., and Wagner, D. (2018).
\newblock {Obfuscated Gradients Give a False Sense of Security: Circumventing
  Defenses to Adversarial Examples}.
\newblock {\em arXiv preprint arXiv:1802.00420}.

\bibitem[Camuto et~al., 2020]{alex2020theoretical}
Camuto, A., Willetts, M., Roberts, S., Holmes, C., and Rainforth, T. (2020).
\newblock {Towards a Theoretical Understanding of The Robustness of Variational
  Autoencoders}.
\newblock {\em arXiv preprint arXiv:2007.07365}.

\bibitem[Cemgil et~al., 2020a]{cemgil2020autoencoding}
Cemgil, T., Ghaisas, S., Dvijotham, K., Gowal, S., and Kohli, P. (2020a).
\newblock {The Autoencoding Variational Autoencoder}.
\newblock In {\em Advances in Neural Information Processing Systems}.

\bibitem[Cemgil et~al., 2020b]{cemgil2020advtraining}
Cemgil, T., Ghaisas, S., Dvijotham, K., and Kohli, P. (2020b).
\newblock {Adversarially Robust Representations with Smooth Encoders}.
\newblock In {\em International Conference on Learning Representations}.

\bibitem[Cohen et~al., 2019]{cohen2019certified}
Cohen, J.~M., Rosenfeld, E., and Kolter, J.~Z. (2019).
\newblock {Certified Adversarial Robustness via Randomized Smoothing}.
\newblock In {\em International Conference on Machine Learning}.

\bibitem[Ghosh et~al., 2019]{Ghosh_2019}
Ghosh, P., Losalka, A., and Black, M.~J. (2019).
\newblock {Resisting Adversarial Attacks using Gaussian Mixture Variational
  Autoencoders}.
\newblock {\em Proceedings of the AAAI Conference on Artificial Intelligence},
  33:541–548.

\bibitem[Ghosh et~al., 2020]{Ghosh2020}
Ghosh, P., Sajjadi, M. S.~M., Vergari, A., Black, M., and Sch{\"{o}}lkopf, B.
  (2020).
\newblock {From Variational to Deterministic Autoencoders}.
\newblock In {\em International Conference on Learning Representations}.

\bibitem[Gondim-Ribeiro et~al., 2018]{gondimribeiro2018adversarial}
Gondim-Ribeiro, G., Tabacof, P., and Valle, E. (2018).
\newblock {Adversarial Attacks on Variational Autoencoders}.
\newblock {\em arXiv preprint arXiv:1806.04646}.

\bibitem[Ha and Schmidhuber, 2018]{ha2018}
Ha, D. and Schmidhuber, J. (2018).
\newblock {World Models}.
\newblock {\em arXiv preprint arXiv:1803.10122}, abs/1803.10122.

\bibitem[Hein and Andriushchenko, 2017]{hein2017formal}
Hein, M. and Andriushchenko, M. (2017).
\newblock {Formal Guarantees on the Robustness of a Classifier Against
  Adversarial Manipulation}.
\newblock In {\em Advances in Neural Information Processing Systems}, pages
  2266--2276.

\bibitem[Higgins et~al., 2017a]{Higgins2017betaVAELB}
Higgins, I., Matthey, L., Pal, A., Burgess, C., Glorot, X., Botvinick, M.,
  Mohamed, S., and Lerchner, A. (2017a).
\newblock {$\beta$-VAE: Learning Basic Visual Concepts with a Constrained
  Variational Framework}.
\newblock In {\em ICLR}.

\bibitem[Higgins et~al., 2017b]{higgins2017darla}
Higgins, I., Pal, A., Rusu, A., Matthey, L., Burgess, C., Pritzel, A.,
  Botvinick, M., Blundell, C., and Lerchner, A. (2017b).
\newblock {Darla: Improving Zero-Shot Transfer in Reinforcement Learning}.
\newblock In {\em International Conference on Machine Learning}, pages
  1480--1490. PMLR.

\bibitem[Huster et~al., 2019]{Huster_2019}
Huster, T., Chiang, C.-Y.~J., and Chadha, R. (2019).
\newblock {Limitations of the Lipschitz Constant as a Defense Against
  Adversarial Examples}.
\newblock {\em Lecture Notes in Computer Science}, page 16–29.

\bibitem[Inglot, 2010]{inglot2010inequalities}
Inglot, T. (2010).
\newblock {Inequalities for Quantiles of the Chi-Square Distribution}.
\newblock {\em Probability and Mathematical Statistics}, 30(2):339--351.

\bibitem[Kim et~al., 2018]{kim2018semiamortized}
Kim, Y., Wiseman, S., Miller, A.~C., Sontag, D., and Rush, A.~M. (2018).
\newblock {Semi-Amortized Variational Autoencoders}.
\newblock {\em arXiv preprint arXiv:1802.02550}.

\bibitem[Kingma and Welling, 2013]{kingma2013autoencoding}
Kingma, D.~P. and Welling, M. (2013).
\newblock {Auto-Encoding Variational Bayes}.
\newblock {\em arXiv preprint arXiv:1312.6114}.

\bibitem[Kingma and Welling, 2019]{Kingma_2019}
Kingma, D.~P. and Welling, M. (2019).
\newblock {An Introduction to Variational Autoencoders}.
\newblock {\em Foundations and Trends® in Machine Learning}, 12(4):307–392.

\bibitem[Kos et~al., 2018]{Kos_2018}
Kos, J., Fischer, I., and Song, D. (2018).
\newblock {Adversarial Examples for Generative Models}.
\newblock {\em 2018 IEEE Security and Privacy Workshops (SPW)}.

\bibitem[Kumar and Poole, 2020]{kumar2020implicit}
Kumar, A. and Poole, B. (2020).
\newblock {On Implicit Regularization in $\beta$-VAEs}.
\newblock {\em arXiv preprint arXiv:2002.00041}.

\bibitem[Li et~al., 2019]{li2019preventing}
Li, Q., Haque, S., Anil, C., Lucas, J., Grosse, R.~B., and Jacobsen, J.-H.
  (2019).
\newblock {Preventing Gradient Attenuation in Lipschitz Constrained
  Convolutional Networks}.
\newblock In {\em Advances in neural information processing systems}, pages
  15390--15402.

\bibitem[Liu et~al., 2009]{liu2009new}
Liu, H., Tang, Y., and Zhang, H.~H. (2009).
\newblock {A New Chi-Square Approximation to the Distribution of Non-Negative
  Definite Quadratic Forms in Non-Central Normal Variables}.
\newblock {\em Computational Statistics \& Data Analysis}, 53(4):853--856.

\bibitem[Loaiza-Ganem and Cunningham, 2019]{loaizaganem2019continuous}
Loaiza-Ganem, G. and Cunningham, J.~P. (2019).
\newblock {The Continuous Bernoulli: Fixing a Pervasive Error in Variational
  Autoencoders}.
\newblock In {\em Advances in Neural Information Processing Systems}, pages
  13287--13297.

\bibitem[Mathieu et~al., 2019]{mathieu2018disentangling}
Mathieu, E., Rainforth, T., Siddharth, N., and Teh, Y.~W. (2019).
\newblock {Disentangling Disentanglement in Variational Autoencoders}.
\newblock In {\em International Conference on Machine Learning}, pages
  4402--4412.

\bibitem[Razavi et~al., 2019]{razavi2019generating}
Razavi, A., van~den Oord, A., and Vinyals, O. (2019).
\newblock {Generating Diverse High-Fidelity Images with VQ-VAE-2}.
\newblock In {\em Advances in Neural Information Processing Systems}, pages
  14866--14876.

\bibitem[Rezende et~al., 2014]{rezende2014stochastic}
Rezende, D.~J., Mohamed, S., and Wierstra, D. (2014).
\newblock {Stochastic Backpropagation and Approximate Inference in Deep
  Generative Models}.
\newblock {\em arXiv preprint arXiv:1401.4082}.

\bibitem[Salman et~al., 2019]{salman2019provably}
Salman, H., Li, J., Razenshteyn, I., Zhang, P., Zhang, H., Bubeck, S., and
  Yang, G. (2019).
\newblock {Provably Robust Deep Learning via Adversarially Trained Smoothed
  Classifiers}.
\newblock In {\em Advances in Neural Information Processing Systems}, pages
  11292--11303.

\bibitem[Schott et~al., 2018]{schott2018adversarially}
Schott, L., Rauber, J., Bethge, M., and Brendel, W. (2018).
\newblock {Towards the First Adversarially Robust Neural Network Model on
  MNIST}.
\newblock {\em arXiv preprint arXiv:1805.09190}.

\bibitem[Shao, 2015]{shao2015}
Shao, J. (2015).
\newblock {Noncentral Chi-Squared, t- and F-Distributions}.
\newblock Lecture.

\bibitem[Szegedy et~al., 2013]{szegedy2013intriguing}
Szegedy, C., Zaremba, W., Sutskever, I., Bruna, J., Erhan, D., Goodfellow, I.,
  and Fergus, R. (2013).
\newblock {Intriguing Properties of Neural Networks}.
\newblock {\em arXiv preprint arXiv:1312.6199}.

\bibitem[Tabacof et~al., 2016]{tabacof2016adversarial}
Tabacof, P., Tavares, J., and Valle, E. (2016).
\newblock {Adversarial Images for Variational Autoencoders}.
\newblock {\em arXiv preprint arXiv:1612.00155}.

\bibitem[Tsuzuku et~al., 2018]{tsuzuku2018lipschitzmargin}
Tsuzuku, Y., Sato, I., and Sugiyama, M. (2018).
\newblock {Lipschitz-Margin Training: Scalable Certification of Perturbation
  Invariance for Deep Neural Networks}.
\newblock In {\em Advances in neural information processing systems}, pages
  6541--6550.

\bibitem[Uesato et~al., 2018]{uesato2018adversarial}
Uesato, J., O'Donoghue, B., Oord, A. v.~d., and Kohli, P. (2018).
\newblock {Adversarial Risk and the Dangers of Evaluating Against Weak
  Attacks}.
\newblock {\em arXiv preprint arXiv:1802.05666}.

\bibitem[Virmaux and Scaman, 2018]{scaman2018lipschitz}
Virmaux, A. and Scaman, K. (2018).
\newblock {Lipschitz Regularity of Deep Neural Networks: Analysis and Efficient
  Estimation}.
\newblock In {\em Advances in Neural Information Processing Systems}, pages
  3835--3844.

\bibitem[Willetts et~al., 2021]{willetts2021improving}
Willetts, M., Camuto, A., Rainforth, T., Roberts, S., and Holmes, C. (2021).
\newblock {Improving VAEs' Robustness to Adversarial Attacks}.
\newblock In {\em International Conference on Learning Representations}.

\bibitem[Yang et~al., 2020]{yang2020}
Yang, Y.-Y., Rashtchian, C., Zhang, H., Salakhutdinov, R., and Chaudhuri, K.
  (2020).
\newblock {Adversarial Robustness Through Local Lipschitzness}.
\newblock {\em arXiv preprint arXiv:2003.02460}.

\end{thebibliography}

\clearpage
\appendix

\thispagestyle{empty}

\onecolumn \makesupplementtitle

\renewcommand\thefigure{\thesection.\arabic{figure}}

\begin{appendices}
\section{PROOFS}
\label{sec: proofs}
\addtocounter{theorem}{-3}

\begin{theorem}[Probability Bound]
Assume $q_\phi(\z|\x)=\gauss{\z}{\mu_\phi(\x)}{\diag{\sigma_\phi^2(\x)}}$ and that the deterministic component of the VAE decoder $g_\theta(\cdot)$ is $a$-Lipschitz, the encoder mean $\mu_\phi(\cdot)$ is $b$-Lipschitz, and the encoder standard deviation $\sigma_\phi(\cdot)$ is $c$-Lipschitz. Finally, let $\pertsample \sim q_\phi(\z|\x+\pert)$ and $\unpertsample \sim q_\phi(\z|\x)$. Then for any $r\in \mathbb{R}^+$, any $\x \in \mathcal{X}$, and any input perturbation $\pert \in \mathcal{X}$,
\[\prob{||g_\theta(\pertsample) - g_\theta(\unpertsample)||_2 \leq r} \geq 1 - \min\left\{p_1(\x), p_2(\x) \right\},\]
where
\[
p_1(\x)\!:=\!\min\left(1, \frac{a^2\left(b^2||\pert||_2^2 + (c||\pert||_2+2||\sigma_\phi(\x)||_2)^2\right)}{r^2}\right)
\]
and
\[
p_2(\x) := \begin{cases}
C(d_z) \frac{u(\x)^{\frac{{d_z}}{2}}\exp\left\{-\frac{u(\x)}{2}\right\}}{u(\x)-{d_z}+2} & \left(\frac{r}{a}-b||\pert||_2\right) \geq 0; {d_z} \geq 2; u(\x) > {d_z}-2\\
1 & \text{o.w.}
\end{cases}
\]
for
$u(\x) := \frac{\left(\frac{r}{a}-b||\pert||_2\right)^2}{\left(c||\pert||_2 + 2||\sigma_\phi(\x)||_2\right)^2}$ and constant $C({d_z}) := \frac{1}{\sqrt{\pi}}\exp\left\{\frac{1}{2}({d_z}-({d_z}-1)\log {d_z})\right\}.$
\end{theorem}

\begin{proof}
Since $g_\theta(\cdot)$ is $a$-Lipschitz,
\begin{equation}
\label{eq: distance_relation}
||g_\theta(\z_1)-g_\theta(\z_2)||_2 \leq a||\z_1-\z_2||_2
\end{equation}
for all $\z_1, \z_2 \in \mathcal{Z}$.

Now assume $\z_1 \sim q_\phi(\z|\x_1)$ and $\z_2 \sim q_\phi(\z|\x_2)$ for some $\x_1, \x_2 \in \mathcal{X}$, such that $g_\theta(\z_1)$ and $g_\theta(\z_2)$ are random variables. \eqref{eq: distance_relation} then implies
\[\left\{||g_\theta(\z_1) - g_\theta(\z_2)||_2 \leq r\right\} \supseteq \left\{a||\z_1-\z_2||_2 \leq r\right\},\]
which in turn implies
\begin{equation}
\label{eq: first_step_extended_proof}
 \prob{||g_\theta(\z_1) - g_\theta(\z_2)||_2 \leq r} \geq \prob{a||\z_1-\z_2||_2 \leq r}.
\end{equation}

Letting $\x_1=\x+\pert$ and $\x_2 = \x$ such that $\z_1=\pertsample$ and $\z_2=\unpertsample$, $q_\phi(\z|\x)=\gauss{\z}{\mu_\phi(\x)}{\diag{\sigma_\phi^2(\x)}}$ means
\[\pertsample \sim q_\phi(\z|\x+\pert)=\mathcal{N}\left(\mu_\phi(\x+\pert), \diag{\sigma_\phi^2(\x+\pert)}\right)\] and
\[\unpertsample \sim q_\phi(\z|\x)=\mathcal{N}\left(\mu_\phi(\x), \diag{\sigma_\phi^2(\x)}\right).\]
Further, since samples from $q_\phi(\z|\cdot)$ are drawn independently in every VAE forward pass, we also know $\pertsample$ and $\unpertsample$ are independent, and thus, because the difference of independent multivariate Gaussian random variables is multivariate Gaussian,
\begin{equation*}
   \pertsample-\unpertsample
   \sim
   \mathcal{N}\left(\mu_\phi(\x+\pert)-\mu_\phi(\x), \diag{\sigma_\phi^2(\x+\pert)} + \diag{\sigma_\phi^2(\x)}\right).
\end{equation*}

Returning to \eqref{eq: first_step_extended_proof}, since $||\pertsample-\unpertsample||_2$ is a continuous random variable, we can write
\begin{equation}
\label{eq: proof_2_starting_point}
    \prob{||g_\theta(\pertsample) - g_\theta(\unpertsample)||_2 \leq r}
     \geq \prob{||\pertsample-\unpertsample||_2 \leq \frac{r}{a}}
     = 1 - \prob{||\pertsample-\unpertsample||_2 \geq \frac{r}{a}}.
\end{equation}
The proof now diverges, yielding $p_1(\x)$ and $p_2(\x)$ respectively.

\paragraph{Obtaining $p_1(\x)$:} Recall $\mathcal{Z}=\mathbb{R}^{d_z}$, apply the definition of the $\ell_2$ norm, and invoke Markov's Inequality to obtain
\begin{equation}
\label{eq: proof_1_markov}
\prob{||\pertsample-\unpertsample||_2 \geq \frac{r}{a}} \\
= \prob{\sum_{j=1}^{d_z} \left(\pertsample-\unpertsample\right)_j^2 \geq \left(\frac{r}{a}\right)^2} \\
\leq \frac{\expec{\sum_{j=1}^{d_z} \left(\pertsample-\unpertsample\right)_j^2}{}}{\left(\frac{r}{a}\right)^2}.
\end{equation}

Now note that
\begin{equation*}
    \sum_{j=1}^{d_z} \left(\pertsample-\unpertsample\right)_j^2
    =\sum_{j=1}^{d_z} \left(\sigma_\phi^2(\x+\pert)+\sigma_\phi^2(\x)\right)_j \frac{\left(\pertsample-\unpertsample\right)_j^2}{\left(\sigma_\phi^2(\x+\pert)+\sigma_\phi^2(\x)\right)_j},
\end{equation*}
so that by the linearity of expectations,
\begin{align}
    & \expec{\sum_{j=1}^{d_z} \left(\pertsample-\unpertsample\right)_j^2}{} \notag \\
    & = \expec{\sum_{j=1}^{d_z} \left(\sigma_\phi^2(\x+\pert)+\sigma_\phi^2(\x)\right)_j \frac{\left(\pertsample-\unpertsample\right)_j^2}{\left(\sigma_\phi^2(\x+\pert)+\sigma_\phi^2(\x)\right)_j}}{} \notag \\
    & \label{eq: full_expectation} = \sum_{j=1}^{d_z} \left(\sigma_\phi^2(\x+\pert)+\sigma_\phi^2(\x)\right)_j \expec{\frac{\left(\pertsample-\unpertsample\right)_j^2}{\left(\sigma_\phi^2(\x+\pert)+\sigma_\phi^2(\x)\right)_j}}{}.
\end{align}

Because $\pertsample-\unpertsample$ is diagonal-covariance multivariate Gaussian, the $\left(\pertsample-\unpertsample\right)_j$ are jointly independent for all $j=1,\ldots, {d_z}$, and so we recognize that
\[\frac{\left(\pertsample-\unpertsample\right)_j^2}{\left(\sigma_\phi^2(\x+\pert)+\sigma_\phi^2(\x)\right)_j}\]
has a non-central $\chi^2$ distribution with one degree of freedom and non-centrality parameter
\[\frac{\left(\mu_\phi(\x + \pert) - \mu_\phi(\x)\right)_j^2}{\left(\sigma_\phi^2(\x + \pert) + \sigma_\phi^2(\x)\right)_j}.\]

Since for a non-central $\chi^2$ random variable $Y$ with $n$ degrees of freedom and non-centrality parameter $\gamma$ \citep{shao2015}, $\expec{Y}{}=n+\gamma$, we have
\[\expec{\frac{\left(\pertsample-\unpertsample\right)_j^2}{\left(\sigma_\phi^2(\x+\pert)+\sigma_\phi^2(\x)\right)_j}}{}=1+\frac{\left(\mu_\phi(\x + \pert) - \mu_\phi(\x)\right)_j^2}{\left(\sigma_\phi^2(\x + \pert) + \sigma_\phi^2(\x)\right)_j},\]
and so plugging into \eqref{eq: full_expectation},
\begin{align*}
    & \expec{\sum_{j=1}^{d_z} \left(\pertsample-\unpertsample\right)_j^2}{} \\
    & = \sum_{j=1}^{d_z} \left(\sigma_\phi^2(\x+\pert)+\sigma_\phi^2(\x)\right)_j \left(1+\frac{\left(\mu_\phi(\x + \pert) - \mu_\phi(\x)\right)_j^2}{\left(\sigma_\phi^2(\x + \pert) + \sigma_\phi^2(\x)\right)_j}\right) \\
    & = \sum_{j=1}^{d_z} \left(\sigma_\phi^2(\x+\pert)+\sigma_\phi^2(\x)\right)_j + \sum_{j=1}^{d_z} \left(\mu_\phi(\x + \pert) - \mu_\phi(\x)\right)_j^2.
\end{align*}

Using
\[\sum_{j=1}^{d_z} \left(\mu_\phi(\x + \pert) - \mu_\phi(\x)\right)_j^2=||\mu_\phi(\x + \pert) - \mu_\phi(\x)||_2^2\]
(the definition of the $\ell_2$ norm), and
\[||\mu_\phi(\x + \pert) - \mu_\phi(\x)||_2 \leq b||\pert||_2,\]
(since $\mu_\phi(\cdot)$ is $b$-Lipschitz), we obtain
\begin{equation}
\label{eq: lip_mean}
  \sum_{j=1}^{d_z} \left(\mu_\phi(\x + \pert) - \mu_\phi(\x)\right)_j^2
 =||\mu_\phi(\x + \pert) - \mu_\phi(\x)||_2^2
 \leq \left(b||\pert||_2\right)^2=b^2||\pert||_2^2.
\end{equation}

Similarly, using
\begin{align*}
& \sum_{j=1}^{d_z} \left(\sigma_\phi^2(\x+\pert)+\sigma_\phi^2(\x)\right)_j \\
& \leq \sum_{j=1}^{d_z} \sigma_\phi^2(\x+\pert)_j+\sigma_\phi^2(\x)_j + 2\sigma_\phi(\x+\pert)_j\sigma_\phi(\x)_j \\
& = \sum_{j=1}^{d_z} \left(\sigma_\phi(\x+\pert)+\sigma_\phi(\x)\right)_j^2 \\
& = \left(\sqrt{\sum_{j=1}^{d_z} \left(\sigma_\phi(\x+\pert)+\sigma_\phi(\x)\right)_j^2}\right)^2 \\
& = ||\sigma_\phi(\x+\pert)+\sigma_\phi(\x)||_2^2
\end{align*}
(where the above inequality follows from $\sigma_\phi: \mathcal{X} \rightarrow \mathbb{R}^{d_z}_{\geq 0}$, and the last equality follows from the definition of the $\ell_2$ norm), and
\begin{align*}
    & ||\sigma_\phi(\x+\pert)+\sigma_\phi(\x)||_2 \\
    & = ||\sigma_\phi(\x+\pert)-\sigma_\phi(\x)+2\sigma_\phi(\x)||_2 \\
    & \leq ||\sigma_\phi(\x+\pert)-\sigma_\phi(\x)||_2+2||\sigma_\phi(\x)||_2 \\
    & \leq c||\pert||_2+2||\sigma_\phi(\x)||_2
\end{align*}
(where the first inequality follows by the triangle inequality, and the second follows from the assumption that $\sigma_\phi(\cdot)$ is $c$-Lipschitz), we find
\begin{equation}
\label{eq: proof_2_key_inequality}
    \sum_{j=1}^{d_z} \left(\sigma_\phi^2(\x+\pert)+\sigma_\phi^2(\x)\right)_j
    \leq ||\sigma_\phi(\x+\pert)+\sigma_\phi(\x)||_2^2
    \leq \left(c||\pert||_2+2||\sigma_\phi(\x)||_2\right)^2.
\end{equation}

Hence, returning to \eqref{eq: proof_1_markov}, we see
\begin{align*}
    & \frac{\expec{\sum_{j=1}^{d_z} \left(\pertsample-\unpertsample\right)_j^2}{}}{\left(\frac{r}{a}\right)^2} \\
    & = \frac{\sum_{j=1}^{d_z} \left(\sigma_\phi^2(\x+\pert)+\sigma_\phi^2(\x)\right)_j + \sum_{j=1}^{d_z} \left(\mu_\phi(\x + \pert) - \mu_\phi(\x)\right)_j^2}{\left(\frac{r}{a}\right)^2} \\
    & \leq \frac{b^2||\pert||_2^2 + \left(c||\pert||_2+2||\sigma_\phi(\x)||_2\right)^2}{\left(\frac{r}{a}\right)^2} \\
    & = \frac{a^2\left(b^2||\pert||_2^2 + (c||\pert||_2+2||\sigma_\phi(\x)||_2)^2\right)}{r^2},
\end{align*}
such that
\begin{equation*}
\prob{||\pertsample-\unpertsample||_2 \geq \frac{r}{a}}
\leq \frac{\expec{\sum_{j=1}^{d_z} \left(\pertsample-\unpertsample\right)_j^2}{}}{\left(\frac{r}{a}\right)^2} \leq \frac{a^2\left(b^2||\pert||_2^2 + (c||\pert||_2+2||\sigma_\phi(\x)||_2)^2\right)}{r^2}.
\end{equation*}
Noting that the right-most term is non-negative, and wanting to have a well-defined probability, we take
\[p_1(\x)\!:=\!\min\left(1, \frac{a^2\left(b^2||\pert||_2^2 + (c||\pert||_2+2||\sigma_\phi(\x)||_2)^2\right)}{r^2}\right),\]
such that
\[\prob{||\pertsample-\unpertsample||_2 \geq \frac{r}{a}} \leq p_1(\x).\]

\paragraph{Obtaining $p_2(\x)$:} Return to \eqref{eq: proof_2_starting_point}. By the triangle inequality,
\begin{equation*}
    ||\pertsample-\unpertsample||_2
    \leq ||\pertsample-\unpertsample - \left(\mu_\phi(\x+\pert)-\mu_\phi(\x)\right)||_2
    + ||\mu_\phi(\x+\pert)-\mu_\phi(\x)||_2,
\end{equation*}
and hence
\begin{align}
     &\quad\ \prob{||\pertsample-\unpertsample||_2 \geq \frac{r}{a}} \label{eq: first_inequality_proof_2} \\
     &\leq \prob{\left(||\pertsample-\unpertsample - \left(\mu_\phi(\x+\pert)-\mu_\phi(\x)\right)||_2
    + ||\mu_\phi(\x+\pert)-\mu_\phi(\x)||_2\right) \geq \frac{r}{a}} \notag \\
     &= \prob{||\pertsample-\unpertsample - \left(\mu_\phi(\x+\pert)-\mu_\phi(\x)\right)||_2
    \geq \left(\frac{r}{a}-||\mu_\phi(\x+\pert)-\mu_\phi(\x)||_2\right)} \notag.
\end{align}

Then, again recalling $\mathcal{Z}=\mathbb{R}^{d_z}$,
\begin{align}
    & \label{eq: proof_2_quantile} \quad\ \prob{||\pertsample-\unpertsample - \left(\mu_\phi(\x+\pert)-\mu_\phi(\x)\right)||_2 \geq \left(\frac{r}{a}-||\mu_\phi(\x+\pert)-\mu_\phi(\x)||_2\right)} \\
    & = \prob{\sum_{j=1}^{d_z} \left(\pertsample-\unpertsample - \left(\mu_\phi(\x+\pert)-\mu_\phi(\x)\right)\right)_j^2 \geq \left(\frac{r}{a}-||\mu_\phi(\x+\pert)-\mu_\phi(\x)||_2\right)^2} \notag \\
    & \label{eq: proof_2_quantiles_start} \leq \prob{\sum_{j=1}^{d_z} \frac{\left(\pertsample-\unpertsample - \left(\mu_\phi(\x+\pert)-\mu_\phi(\x)\right)\right)_j^2}{\left(\sigma_\phi^2(\x + \pert) + \sigma_\phi^2(\x)\right)_j} \geq \frac{\left(\frac{r}{a}-||\mu_\phi(\x+\pert)-\mu_\phi(\x)||_2\right)^2}{\left(c||\pert||_2+2||\sigma_\phi(\x)||_2\right)^2}},
\end{align}
where the first equality uses the definition of the $\ell_2$ norm, and the above inequality between probabilities uses the inequality from \eqref{eq: proof_2_key_inequality}.

Now, since
\begin{equation*}
    \pertsample-\unpertsample \sim \mathcal{N}\left(\mu_\phi(\x+\pert)-\mu_\phi(\x), \diag{\sigma_\phi^2(\x+\pert)} + \diag{\sigma_\phi^2(\x)} \right),
\end{equation*}
it follows that
\[\frac{\left(\pertsample-\unpertsample - (\mu_\phi(\x+\pert)-\mu_\phi(\x))\right)_j}{\sqrt{\left(\sigma_\phi^2(\x + \pert) + \sigma_\phi^2(\x)\right)_j}} \sim \mathcal{N}(0, 1).\]
In particular, note that since $\pertsample-\unpertsample$ is diagonal-covariance multivariate Gaussian, the
\[\frac{\left(\pertsample-\unpertsample - (\mu_\phi(\x+\pert)-\mu_\phi(\x))\right)_j}{\sqrt{\left(\sigma_\phi^2(\x + \pert) + \sigma_\phi^2(\x)\right)_j}}\]
are jointly independent for all $j=1, \ldots, d_z$. Hence, because the sum of squares of $d_z$ independent standard Gaussian random variables has a standard $\chi^2$ distribution with $d_z$ degrees of freedom,
\[\sum_{j=1}^{d_z} \frac{\left(\pertsample-\unpertsample - (\mu_\phi(\x+\pert)-\mu_\phi(\x))\right)_j^2}{\left(\sigma_\phi^2(\x + \pert) + \sigma_\phi^2(\x)\right)_j} =: Y \sim \chi^2_{d_z}.\]

Letting
\[u'(\x) := \frac{\left(\frac{r}{a}-||\mu_\phi(\x+\pert)-\mu_\phi(\x)||_2\right)^2}{\left(c||\pert||_2+2||\sigma_\phi(\x)||_2\right)^2} \quad \text{and} \quad u(\x) := \frac{\left(\frac{r}{a}-b||\pert||_2\right)^2}{\left(c||\pert||_2+2||\sigma_\phi(\x)||_2\right)^2},\]
we have $u'(\x) \geq u(\x)$ by the assumption that $\mu_\phi(\cdot)$ is $b$-Lipschitz, since
\[||\mu_\phi(\x+\pert)-\mu_\phi(\x)||_2 \leq b||\pert||_2,\]
and therefore
\[\left(\frac{r}{a}-||\mu_\phi(\x+\pert)-\mu_\phi(\x)||_2\right) \geq \left(\frac{r}{a}-b||\pert||_2\right)\]
(note also that $\left(c||\pert||_2+2||\sigma_\phi(\x)||_2\right)^2 \geq 0$). Then, using \eqref{eq: proof_2_quantiles_start} with the requirement that
\[\left(\frac{r}{a}-||\mu_\phi(\x+\pert)-\mu_\phi(\x)||_2\right) \geq \left(\frac{r}{a}-b||\pert||_2\right) \geq 0\]
to ensure the inequality in \eqref{eq: proof_2_quantile} is meaningful,
\[\prob{Y \geq u'(\x)} \leq \prob{Y \geq u(\x)}.\]
The tail bound for standard $\chi^2$ random variables in (3.1) from \cite{inglot2010inequalities} (which requires $u(\x) > d_z-2$ and $d_z\geq 2$) then yields
\[\prob{Y \geq u(\x)} \leq C(d_z) \frac{u(\x)^{\frac{d_z}{2}}\exp\left\{-\frac{u(\x)}{2}\right\}}{u(\x)-d_z+2}\]
for constant $C(d_z) :=\frac{1}{\sqrt{\pi}}\exp\left\{\frac{1}{2}(d_z-(d_z-1)\log d_z)\right\}$. Since the expression on the right-hand side is non-negative under the above conditions, we define
\[p_2(\x) := \begin{cases}
C(d_z) \frac{u(\x)^{\frac{{d_z}}{2}}\exp\left\{-\frac{u(\x)}{2}\right\}}{u(\x)-{d_z}+2} & \left(\frac{r}{a}-b||\pert||_2\right) \geq 0; {d_z} \geq 2; u(\x) > {d_z}-2\\
1 & \text{o.w.}
\end{cases}\]
to ensure a well-defined probability. Then, by the inequalities starting from \eqref{eq: first_inequality_proof_2},
\[\prob{||\pertsample-\unpertsample||_2 \geq \frac{r}{a}} \leq p_2(\x).\]

\paragraph{Obtaining the final bound:} Choosing the least of $p_1(\x)$ and $p_2(\x)$ to obtain the tighter upper bound on $\prob{||\pertsample-\unpertsample||_2 \geq \frac{r}{a}}$, we can plug in to \eqref{eq: proof_2_starting_point}, which gives
\begin{align*}
    & \prob{||g_\theta(\pertsample) - g_\theta(\unpertsample)||_2 \leq r} \\
    & \geq 1 - \prob{||\pertsample-\unpertsample||_2 \geq \frac{r}{a}} \\
    & \geq 1 - \min\{p_1(\x), p_2(\x)\}.
\end{align*}
\end{proof}

\bigskip
\bigskip

\begin{lemma}[Margin Bound]
Given the assumptions of Theorem \ref{thm: probability_bound} and some $\epsilon \in [0, 1)$, the $(r, \epsilon)$-robustness margin of this VAE on input $\x$,
\[R^{(r, \epsilon)}(\x) \geq \max \left\{m_1(\x), m_2(\x) \right\}\]
for
\[m_1(\x) := \frac{-4c||\sigma_\phi(\x)||_2 + \sqrt{\left(4c||\sigma_\phi(\x)||_2\right)^2
    -4\left(c^2+b^2\right)\left(4||\sigma_\phi(\x)||_2 - (1-\epsilon) \left(\frac{r}{a}\right)^2\right)}}{2\left(c^2+b^2\right)}\]
and $m_2(\x) := \sup \left\{||\pert||_2 : p_2(\pert, \x) \leq (1-\epsilon) \right\},$
where $p_2(\pert, \x)$ is as in Theorem \ref{thm: probability_bound} and we augment the listed arguments of $p_2$ to make explicit the dependence on $\pert$.
\end{lemma}

\begin{proof}
By Theorem \ref{thm: probability_bound}, for any input perturbation $\pert \in \mathcal{X}$ and any input $\x \in \mathcal{X}$,
\[\prob{||g_\theta(\pertsample) - g_\theta(\unpertsample)||_2 \leq r} \geq 1 - \min\{p_1(\x), p_2(\x)\}.\]
Hence, for our Lipschitz-VAE to be $(r, \epsilon)$-robust to perturbation $\pert$ on input $\x$ for threshold $\epsilon \in [0, 1)$, by Definition \ref{def: r_robust} it suffices that
\[1 - \min\{p_1(\x), p_2(\x)\} > \epsilon.\]
Recalling Definition \ref{def: robustness_margin}, since for a model $f$, $R^{(r, \epsilon)}(\x)$ is defined by
\[||\pert||_2 < R^{(r, \epsilon)}(\x) \implies \prob{||f(\x+\pert)-f(\x)||_2 \leq r} > \epsilon,\]
for our Lipschitz-VAE $R^{(r, \epsilon)}(\x)$ is at least the maximum perturbation norm such that
\[1 - \min\{p_1(\pert, \x), p_2(\pert, \x)\} \geq \epsilon,\]
or equivalently,
\begin{equation*}
\label{eq: margin_inequality}
    \max \left\{\sup \left\{ ||\pert||_2: p_1(\pert, \x) \leq (1-\epsilon)\right\},\ \sup\left\{||\pert||_2: p_2(\pert, \x) \leq (1-\epsilon) \right\} \right\}
\end{equation*}
(where we make explicit the dependence on $\pert$).

Denoting $m_1(\x) := \sup\left\{ ||\pert||_2: p_1(\pert, \x) \leq (1-\epsilon)\right\}$ and rearranging, $m_1(\x)$ becomes
\[\sup\left\{ ||\pert||_2: \left(c^2+b^2\right)||\pert||_2^2 + 4c||\sigma_\phi(\x)||_2||\pert||_2 + 4||\sigma_\phi(\x)||_2^2-(1-\epsilon)\left(\frac{r}{a}\right)^2 \leq 0\right\}.\]
Excluding the degenerate case of $c = 0$, that is assuming $c > 0$, this is attained at the maximum root of the quadratic equation
\[\left(c^2+b^2\right)||\pert||_2^2 + 4c||\sigma_\phi(\x)||_2||\pert||_2 + 4||\sigma_\phi(\x)||_2^2-(1-\epsilon)\left(\frac{r}{a}\right)^2 = 0,\]
provided a root exists, and so by the quadratic formula,
\[m_1(\x) = \frac{-4c||\sigma_\phi(\x)||_2 + \sqrt{\left(4c||\sigma_\phi(\x)||_2\right)^2-4\left(c^2+b^2\right)\left(4||\sigma_\phi(\x)||_2 - (1-\epsilon)\left(\frac{r}{a}\right)^2\right)}}{2\left(c^2+b^2\right)}.\]
The second case does not admit a closed-form solution, so we will simply write
\[m_2(\x) :=\sup \left\{||\pert||_2 : p_2(\pert, \x) \leq (1-\epsilon)\right\}.\]
Choosing the maximum of $m_1(\x)$ and $m_2(\x)$ then yields
\[R^{(r, \epsilon)}(\x) \geq \max \left\{m_1(\x), m_2(\x) \right\}.\]
\end{proof}

\bigskip
\bigskip

\begin{theorem}
[Global Margin Bound]
Given the assumptions of Lemma \ref{lem: margin_bound}, but with $\sigma_\phi(\x)=\boldsymbol{\sigma} \in \mathbb{R}^{d_z}_{\geq 0}$, the $(r, \epsilon)$-robustness margin of this VAE for all inputs is
\[R^{(r, \epsilon)} \geq \max \left\{m_1, m_2 \right\},\]
where
\[m_1 := \frac{\sqrt{-\left(4||\boldsymbol{\sigma}||_2^2-(1-\epsilon)\left(\frac{r}{a}\right)^2 \right)}}{b}\]
and $m_2 := \sup \left\{||\pert||_2 : p_2(\pert) \leq (1-\epsilon)\right\}$,
where $p_2$ is as in Theorem \ref{thm: probability_bound}, but $u := \frac{\left(\frac{r}{a}-b||\pert||_2\right)^2}{4||\boldsymbol{\sigma}||_2^2}.$
\end{theorem}

\begin{proof}
Given a fixed encoder standard deviation, that is substituting $\sigma_\phi(\x)=\boldsymbol{\sigma} \in \mathbb{R}^{d_z}_{\geq 0}$, we first have to derive a lower bound on the $r$-robustness probability to then bound the $(r, \epsilon)$-robustness margin globally. We do this using the machinery of Theorem \ref{thm: probability_bound}, which — lifting the now-redundant requirement that the encoder standard deviation be $c$-Lipschitz — can be invoked without loss of generality.

In the case of $p_1$ (recall the two bounds in the proof of Theorem \ref{thm: probability_bound}), plugging in $\boldsymbol{\sigma}$ yields
\begin{align*}
\prob{||g_\theta(\pertsample) - g_\theta(\unpertsample)||_2 \leq r}
    & \geq 1 - \frac{\expec{\sum_{j=1}^{d_z} \left(\pertsample-\unpertsample\right)_j^2}{}}{\left(\frac{r}{a}\right)^2} \\
    & = 1 - \frac{\sum_{j=1}^{d_z} \left(\boldsymbol{\sigma}^2+\boldsymbol{\sigma}^2\right)_j + \sum_{j=1}^{d_z} \left(\mu_\phi(\x + \pert) - \mu_\phi(\x)\right)_j^2}{(\frac{r}{a})^2} \\
    & \geq 1 - \frac{b^2||\pert||_2^2 + 4||\boldsymbol{\sigma}||_2^2}{\left(\frac{r}{a}\right)^2} \\
    & = 1 - p_1
\end{align*}
for $p_1 :=  \frac{a^2\left(b^2||\pert||_2^2 + 4||\boldsymbol{\sigma}||_2^2\right)}{r^2}$, where the penultimate step follows by \eqref{eq: lip_mean} and \eqref{eq: proof_2_key_inequality}. In the case of $p_2$, we can directly substitute, obtaining
\[ \prob{||g_\theta(\pertsample) - g_\theta(\unpertsample)||_2 \leq r} \geq 1 - p_2 \]
for
\[p_2 := \begin{cases}
C(d_z) \frac{u^{\frac{{d_z}}{2}}\exp\left\{-\frac{u}{2}\right\}}{u-{d_z}+2} & \left(\frac{r}{a}-b||\pert||_2\right) \geq 0; {d_z} \geq 2; u > {d_z}-2\\
1 & \text{o.w.}
\end{cases}\]
and
$u := \frac{\left(\frac{r}{a}-b||\pert||_2\right)^2}{4||\boldsymbol{\sigma}||_2^2}$. Theorem \ref{thm: glob_stoch_margin_bound} then follows by identical reasoning to Lemma \ref{lem: margin_bound}.
\end{proof}

\section{IMPLEMENTING CERTIFIABLY ROBUST VAES}
\label{app:lipschitz_implement}

Ensuring the Lipschitz continuity of a deep learning architecture is non-trivial in practice. Using \cite{anil2018sorting} as a guide, this section elaborates on how to provably control the Lipschitz constants of an encoder and decoder network.\footnote{For simplicity, we focus on fully-connected architectures, although the same ideas extend, for example, to convolutional architectures \citep{li2019preventing}.}

We define a fully-connected network with $L$ layers as the composition of linear transformations $\mathbf{W}_l$ and element-wise activation functions $\varphi_l(\cdot)$ for $l=1, \ldots, L$, where the output of the $l$-th layer
\[\mathbf{h}_l := \varphi_l(\mathbf{W}_{l}\mathbf{h}_{l-1}).\]

\subsection{Ensuring Lipschitz Continuity With Constant $1$}
We would like to ensure a fully-connected network is $M$-Lipschitz for arbitrary Lipschitz constant $M$.
It has been shown that a natural way to achieve this is by first requiring Lipschitz continuity with constant $1$~\citep{anil2018sorting}.

As $1$-Lipschitz functions are closed under composition, if we can ensure that for every layer $l$, $\mathbf{W}_l$ and $\varphi_l(\cdot)$ are $1$-Lipschitz, then the entire network will be $1$-Lipschitz. Most commonly-used activation functions, such as the ReLU and the sigmoid function, are already $1$-Lipschitz \citep{Huster_2019, scaman2018lipschitz}, and hence we need only ensure that $\mathbf{W}_l$ is also $1$-Lipschitz.

This can be done by requiring $\mathbf{W}_l$ to be orthonormal, since $\mathbf{W}_l$ being $1$-Lipschitz is equivalent to the condition
\begin{equation}
\label{eq: condition_linear_map}
    \sup_{||\x||_2 \leq 1}||\mathbf{W}_l\x||_2 \leq 1,
\end{equation}
where $\sup_{||\x||_2 \leq 1}||\mathbf{W}_l\x||_2$ equals the largest singular value of $\mathbf{W}_l$. The singular values of an orthonormal matrix all equal $1$, and so the orthonormality of $\mathbf{W}_l$ implies \eqref{eq: condition_linear_map} is satisfied.

In practice, $\mathbf{W}_l$ can be made orthonormal through an iterative algorithm called \emph{Bj\"{o}rck Orthonormalization}, which on input matrix $\mathbf{A}$ finds the ``nearest'' orthonormal matrix to $\mathbf{A}$ \citep{anil2018sorting}. Bj\"{o}rck Orthonormalization is differentiable and so allows the encoder and decoder networks of a Lipschitz-VAE to be trained using gradient-based methods, just like a standard VAE.

\subsection{Ensuring Lipschitz Continuity With Arbitrary Constants}
Now that we can train a $1$-Lipschitz network, we would like to generalize this method to arbitrary Lipschitz constant $M$. To do so, note that if layer $l$ has Lipschitz constant $M_l$, then the Lipschitz constant of the entire network is $M=\prod_{l=1}^L M_l$ \citep{szegedy2013intriguing}.

Hence, for our $L$-layer fully-connected neural network to be $M$-Lipschitz, it suffices to ensure that each layer $l$ has Lipschitz constant $M^{\frac{1}{L}}$. This is actually simple to achieve, because if we continue to assume $\varphi_l(\cdot)$ is $1$-Lipschitz, a Lipschitz constant of $M^{\frac{1}{L}}$ in layer $l$ follows from scaling the outputs of each layer by $M^{\frac{1}{L}}$.

\subsection{Selecting Activation Functions}
\label{sec: activation_functions}
While the above approach is sufficient to train networks with arbitrary Lipschitz constants, a result from \cite{anil2018sorting} shows it is not sufficient to ensure the resulting networks are also expressive in the space of Lipschitz continuous functions.
Informally, the result states that the expressivity of a Lipschitz-constrained network is limited when its activation functions are not gradient norm-preserving. Since non-linearities such as the ReLU and the sigmoid function do not preserve the gradient norm, the expressivity of Lipschitz-constrained networks that use such activations will be further limited.

To address this, \cite{anil2018sorting} introduces a gradient norm-preserving activation function called \emph{GroupSort}, which in each layer $l$ groups the entries of matrix-vector product $\mathbf{W}_l\mathbf{h}_{l-1}$ into some number of groups, and then sorts the entries of each group by ascending order. It can be shown that when each group has size two,
\[\begin{pmatrix}
1 & 0
\end{pmatrix}^\intercal \text{GroupSort}\left(\begin{pmatrix}
y \\
0
\end{pmatrix}\right)=\text{ReLU}(y)\]
for any scalar $y$ \citep{anil2018sorting}.
Unless we need to restrict a network's outputs to a specific range, we employ the GroupSort activation in our implementation of Lipschitz-VAEs.

\begin{figure*}
\captionsetup[subfloat]{farskip=-1pt,captionskip=-1pt}
\centering
    \begin{tabular}{c c}
        \subfloat[Standard VAE]{\includegraphics[width=0.33\textwidth]{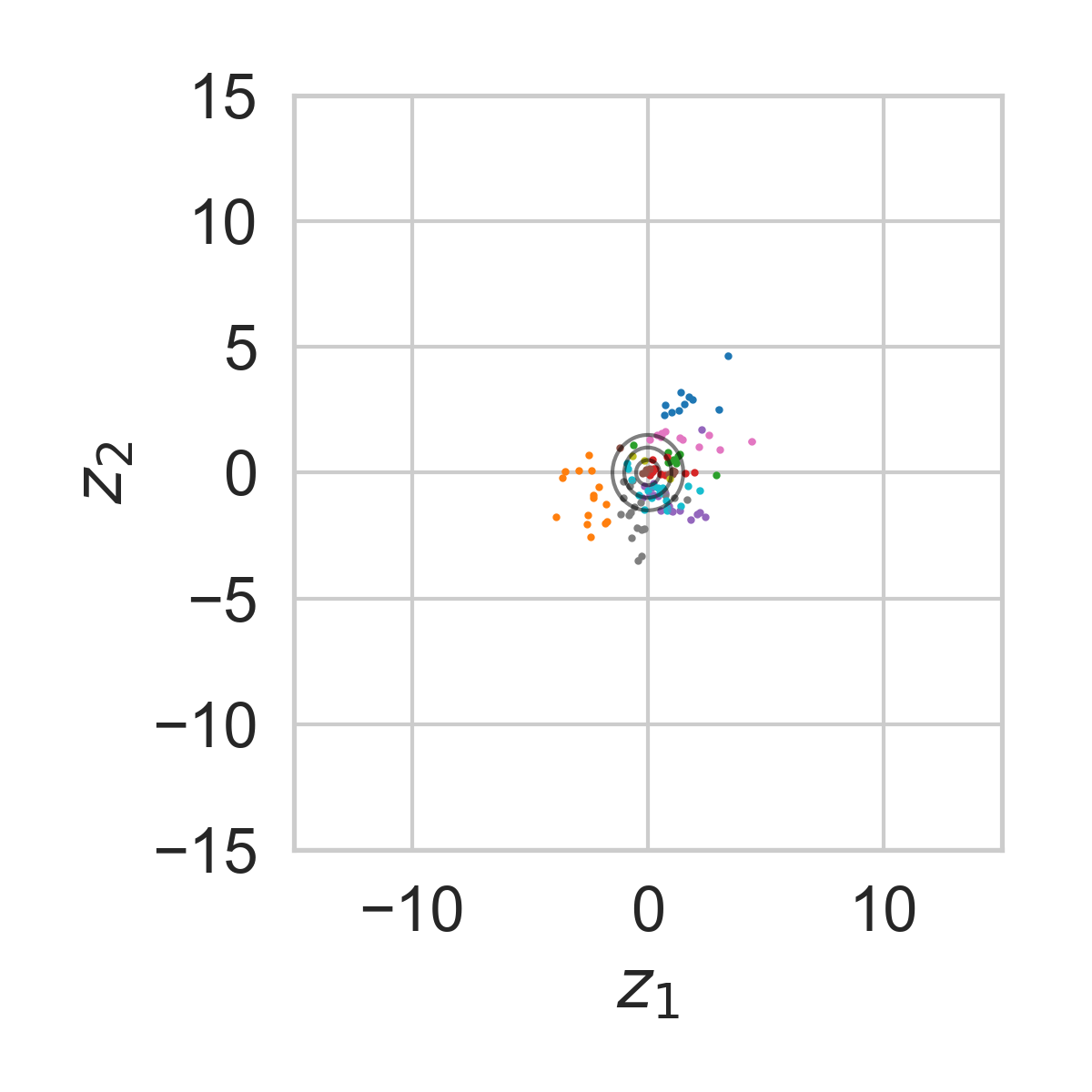}} & \\ \subfloat[Lipschitz-VAE, $M = 5$]{\includegraphics[width=0.33\textwidth]{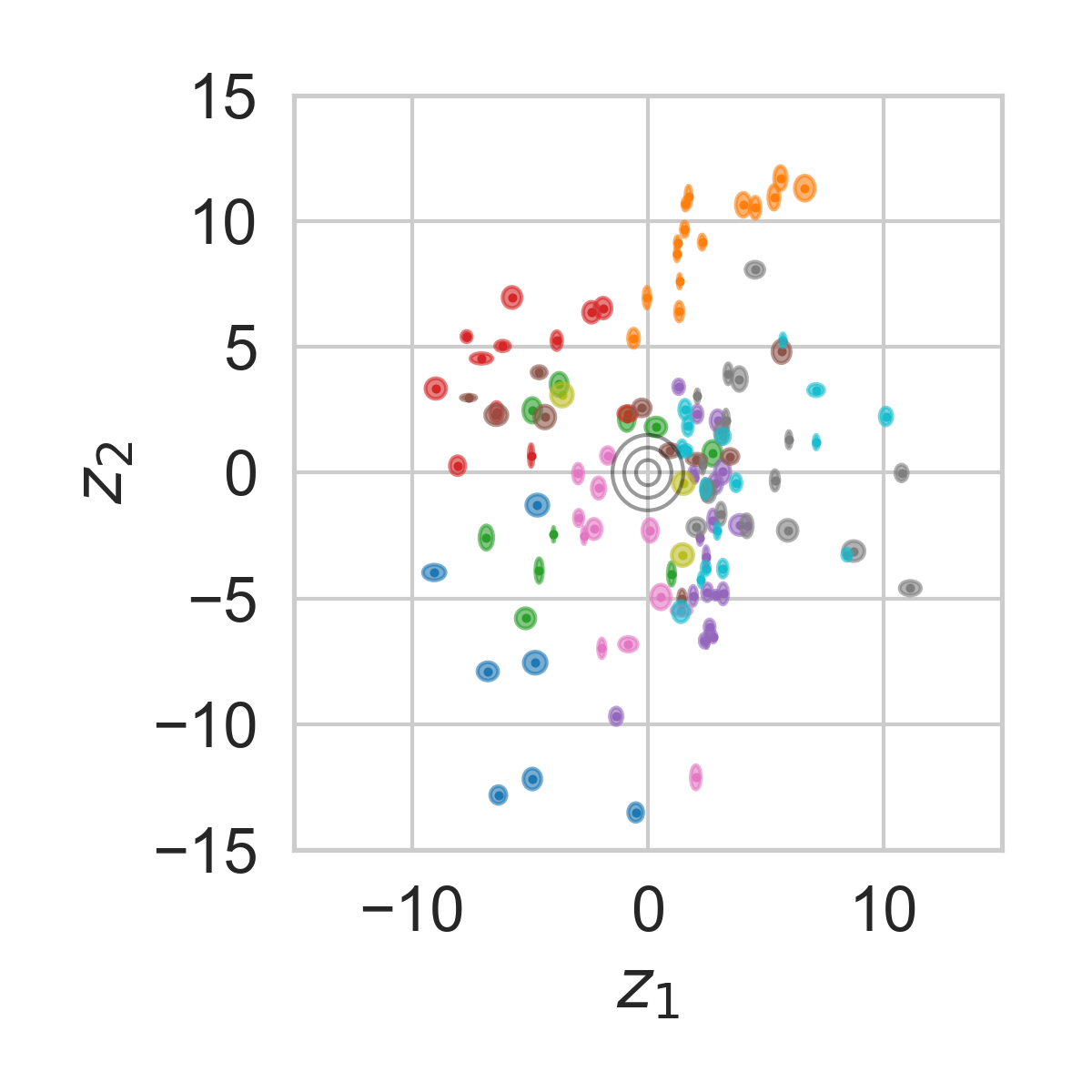}} & \subfloat[Lipschitz-$\beta$-VAE, $M = 5$, $\beta=5$]{\includegraphics[width=0.33\textwidth]{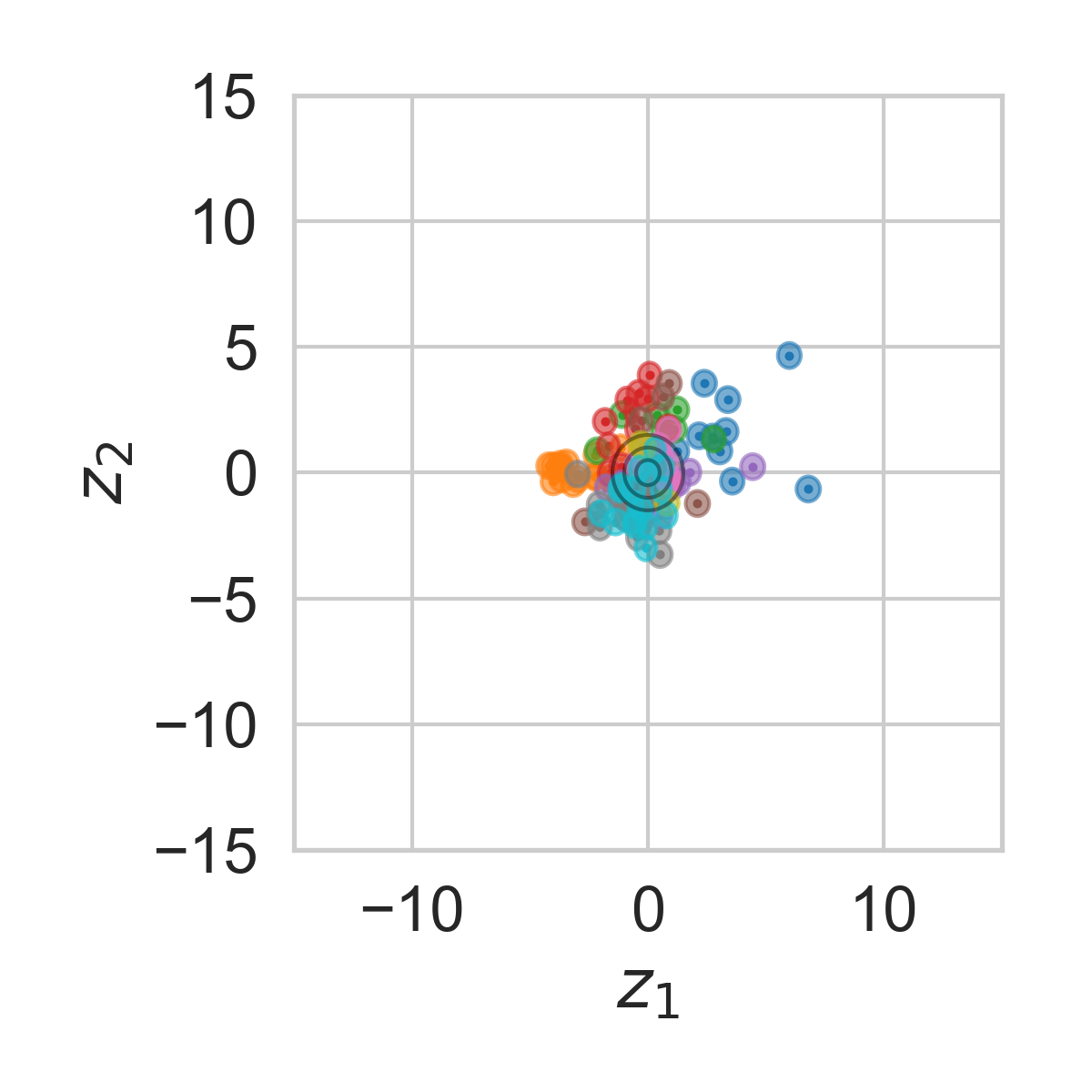}}\\ \subfloat[Lipschitz-VAE, $M = 10$]{\includegraphics[width=0.33\textwidth]{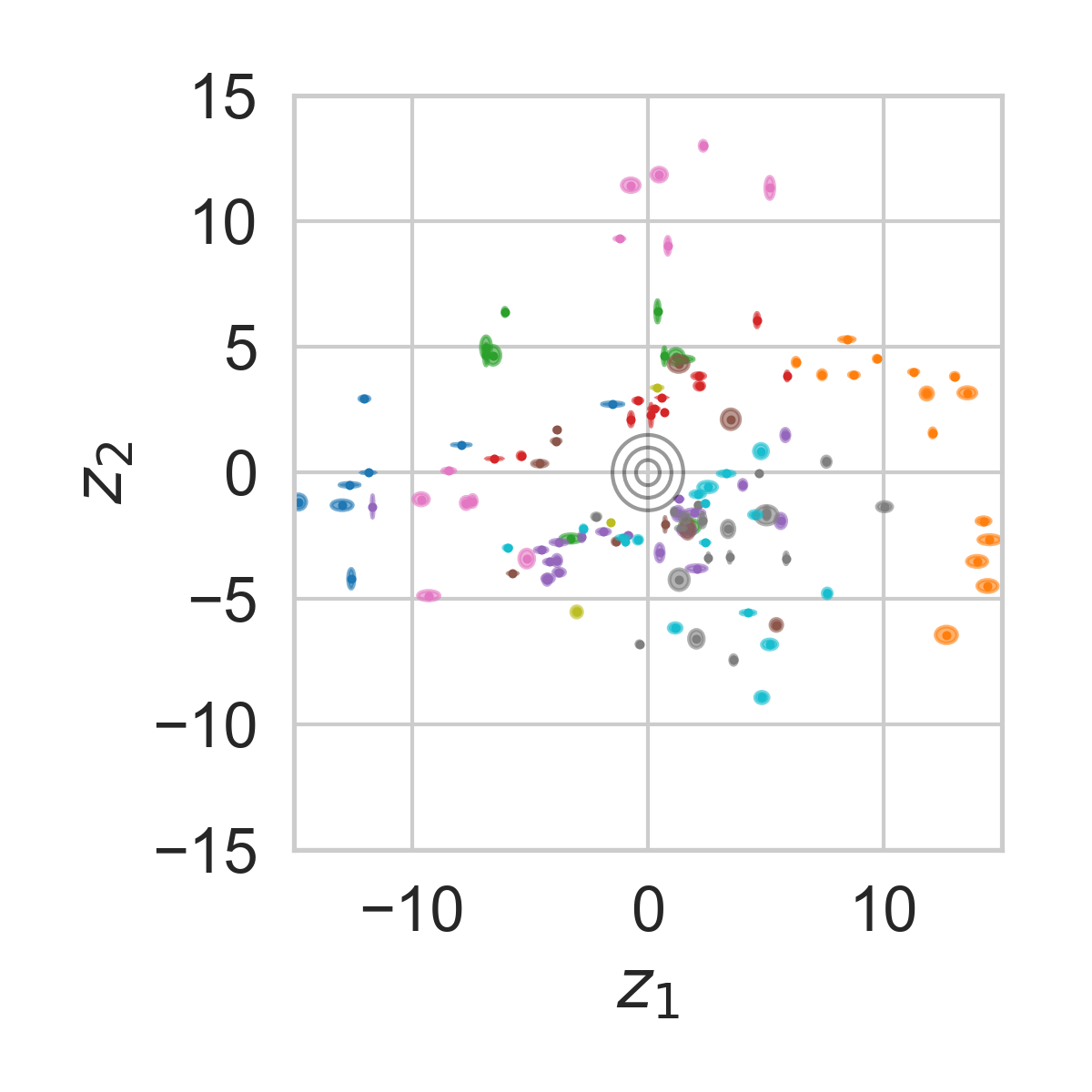}} & \subfloat[Lipschitz-$\beta$-VAE, $M = 10$, $\beta=5$]{\includegraphics[width=0.33\textwidth]{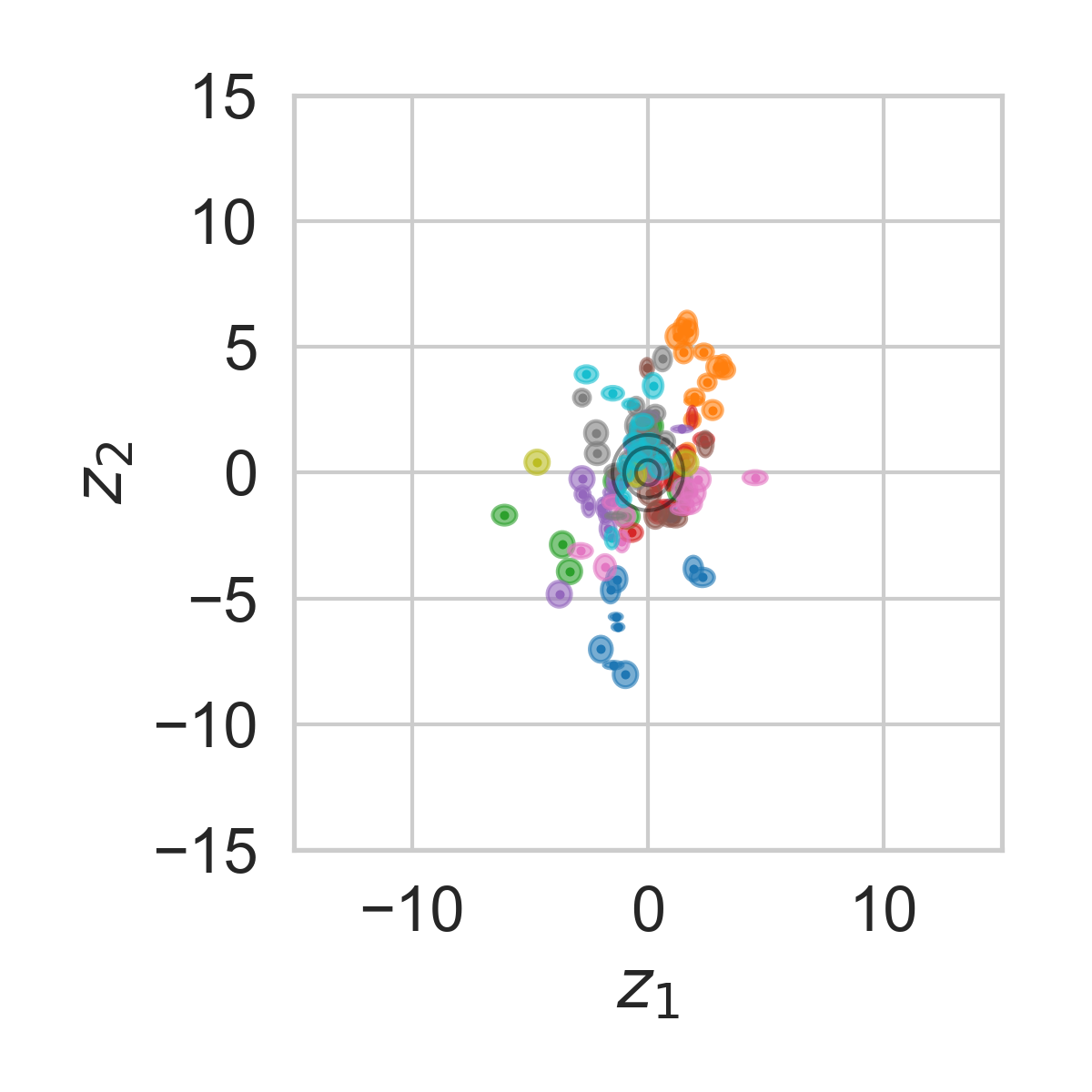}}\\
    \end{tabular}
    \caption{Learned encodings for different types of VAE on MNIST. A colored ellipse represents the posterior $q_\phi(\z|\x_i)$ for a single $\x_i$. The prior, $p(\z)=\mathcal{N}({\z};{\mathbf{0}},{\mathbf{I}})$, is overlaid in black for one, two and three standard deviations. Lipschitz-VAEs have encoders that are dispersed in latent space, in contrast with the learned encoder of a standard VAE. Upweighting the KL term in \eqref{eq: likelihood_lower_bound}, as in a $\beta$-VAE \citep{Higgins2017betaVAELB}, changes this behaviour.}
    \label{fig: learned_latent_spaces_app}
\end{figure*}

\section{INVESTIGATING LEARNED LATENT SPACES}

\label{app:latent_space_explore}
While we are primarily interested in the robustness of VAEs, and their certifiably robust instantiation in Lipschitz-VAEs, we may also wish to understand whether Lipschitz-VAEs qualitatively differ from standard VAEs.

To build our understanding in this regard, we study the latent spaces learned by Lipschitz-VAEs, training standard and Lipschitz-VAEs with latent space dimension $d_z=2$ and visualizing their learned encoders $q_\phi(\z|\x)$. As shown in Figure \ref{fig: learned_latent_spaces_app}, the encoders learned by Lipschitz- and standard VAEs differ in their scale. Whereas the encoder of a standard VAE remains tightly clustered about the prior $p(\z)=\gauss{\z}{\mathbf{0}}{\mathbf{I}}$, the encoders of the Lipschitz-VAEs disperse mass widely in latent space.

This apparent rescaling of the latent space in Lipschitz-VAEs has two important consequences, the first of which is that the prior and encoder have little overlap. This is significant because it is common to generate data points with a trained VAE by drawing samples from the prior and passing these to the decoder. In a rescaled latent space where the prior and encoder have little overlap, many samples from the prior will be ``out-of-distribution'' inputs to the decoder.

The second consequence of the latent space being rescaled is that there risks being less overlap between $q_\phi(\z|\cdot)$ for any two inputs. In the limit, the latent space then devolves into a look-up table \citep{mathieu2018disentangling}, which is undesirable because the meaning of interpolated points in latent space — that is, points between areas of high density in terms of $q_\phi(\z|\cdot)$ — is lost.

We speculate that the rescaling of latent spaces in Lipschitz-VAEs can be explained by the relative importance of the likelihood and KL terms, $\log p_\theta(\x|\z)$ and $\kl{q_\phi(\z|\x)}{p(\z)}$ respectively, in the VAE objective in \eqref{eq: likelihood_lower_bound}. By Definition \ref{def: lipschitz_continuity}, a Lipschitz continuous function is one whose rate of change is constrained, so in some sense such a function is ``simpler'' than others not satisfying the property. It seems plausible then that — to achieve good input reconstructions while using simpler functions than a standard VAE — a Lipschitz-VAE might rescale the latent space to be able to adequately differentiate between latent samples corresponding to different inputs. This might happen even at the expense of the encoder being distant from the prior, causing $\kl{q_\phi(\z|\x)}{p(\z)}$ to grow, since the likelihood term typically dominates the KL term and so gains in the likelihood term from rescaling the latent space might outweigh the resulting penalty from the KL term.

We test this hypothesis by training Lipschitz-VAEs with the KL term upweighted by hyperparameter $\beta$, as in a $\beta$-VAE \citep{Higgins2017betaVAELB} (we term Lipschitz-VAEs trained with this modified objective \emph{Lipschitz-$\beta$-VAEs}). As can be seen in Figure \ref{fig: learned_latent_spaces_app}, and as predicted by our hypothesis, we find that by increasing the weight assigned to the KL term — that is, using $\beta > 1$ — the scaling of the latent space is mitigated.

In sum, the experiments in this section reveal that Lipschitz-VAEs learn qualitatively different encoders from standard VAEs, exhibiting rescaling behavior that we link both to the challenge of performing reconstructions using Lipschitz continuous functions and the characteristics of the VAE objective. Our experiments also outline how possible adverse effects of Lipschitz continuity constraints on data generation and latent space interpretability might be addressed through a small modification of the VAE objective.

\section{ESTIMATING THE $(r, \epsilon)$-ROBUSTNESS MARGIN}
\label{sec:app:est_margin}

\begin{algorithm}
\SetKwInOut{Inputs}{Inputs}
\SetKwInOut{Output}{Output}
\SetKwFunction{MaxDamage}{MaxDamageAttack}
\SetKwBlock{Estimation}{Estimation routine}{Estimation routine}
\Inputs{$\x$, $r$, $\epsilon$, starting estimate \texttt{max\_R}, step size $\alpha$, number of samples $S$, number of random restarts $T$}
\Output{Estimated $(r, \epsilon)$-robustness margin $\hat{R}^{(r, \epsilon)}(\x)$}
\nonl
\Estimation{
$\hat{R}^{(r, \epsilon)}(\x) \leftarrow$ \texttt{max\_R}\;

\nonl
\While{$\hat{R}^{(r, \epsilon)}(\x) > 0$}{
probabilities $\leftarrow []$\;

\nonl
\For{$t=1, \ldots, T$}{
\tcp{Performs a maximum damage attack according to the objective in (\ref{eq: max_damage_attack})}
$\pert_t \leftarrow$ \MaxDamage with the constraint $||\pert||_2 \leq \hat{R}^{(r, \epsilon)}(\x)$; randomly initialized\;

distances $\leftarrow []$\;

\For{$s=1, \ldots, S$}{
    $\z_{\pert_t} \sim q_\phi(\z|\x+\pert_t)$\;
    $\z_{\neg \pert_t} \sim q_\phi(\z|\x)$\;

    distances.append($||g_\theta(\z_{\pert_t})-g_\theta(\z_{\neg \pert_t})||_2$)
}
\tcp{Estimates the $r$-robustness probability}
probability $\leftarrow \frac{\text{length(distances[distances $\leq r$])}}{S}$\;

probabilities.append(probability)\;
}
\tcp{Checks that the estimated probabilities are greater than $\epsilon$, across random restarts}
\If{\upshape length(probabilities[probabilities $> \epsilon$])$=T$}{
\KwRet{$\hat{R}^{(r, \epsilon)}(\x)$}
}
$\hat{R}^{(r, \epsilon)}(\x) \leftarrow \hat{R}^{(r, \epsilon)}(\x)-\alpha $
}
\tcp{Indicates when no positive $(r, \epsilon)$-robustness margin is found}
\KwRet{``No positive $R^{(r, \epsilon)}(\x)$ found.''}}
\caption{\cite{alex2020theoretical}'s algorithm to estimate $(r, \epsilon)$-robustness margin $R^{(r, \epsilon)}(\x)$. Starting with estimate $\texttt{max\_R}$ and decrementing by step size $\alpha$ at each iteration (until reaching $0$), the algorithm performs $T$ maximum damage attacks with input perturbations constrained to the current estimate for the $(r, \epsilon)$-robustness margin. The first time $(r, \epsilon)$-robustness is satisfied under all $T$ attacks, the algorithm returns the current estimate as the estimated $(r, \epsilon)$-robustness margin $\hat{R}^{(r, \epsilon)}(\x)$.}
\label{alg: R_estimation_algorithm}
\end{algorithm}

\newpage
\section{QUALITATIVELY EVALUATING ROBUSTNESS}
\label{app:qualitative_eval}
\begin{figure*}[h!]
    \captionsetup[subfloat]{farskip=-1pt,captionskip=-1pt}
    \begin{tabular}{c c}
        \subfloat[Standard VAE, $||\pert||_2 \leq 1$.]{\includegraphics[width=0.49\textwidth]{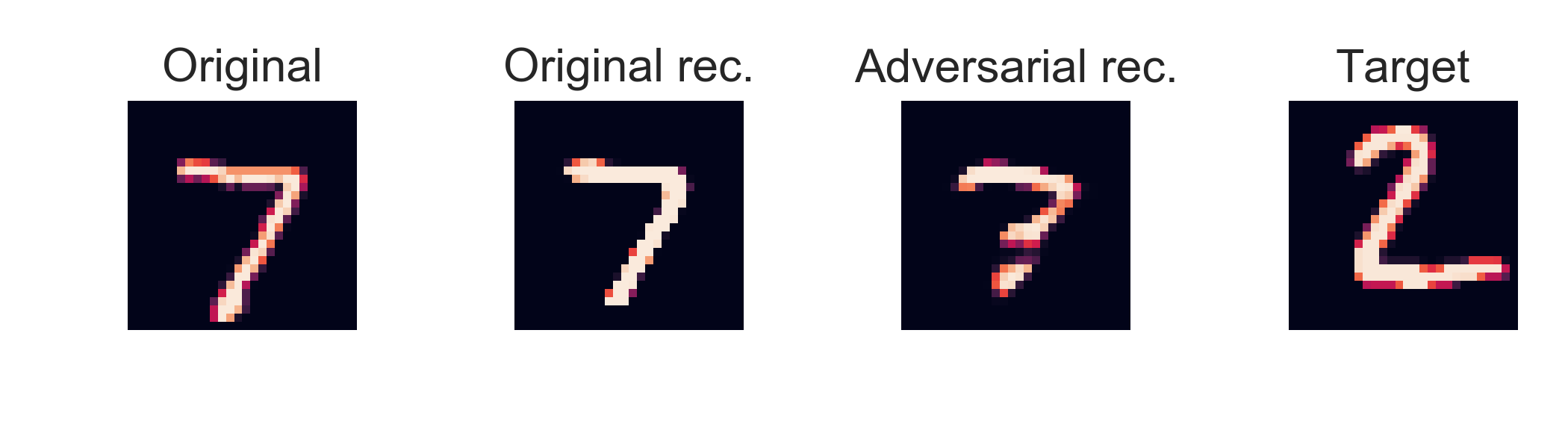}} & \subfloat[Lipschitz-VAE, $||\pert||_2 \leq 1$.]{\includegraphics[width=0.49\textwidth]{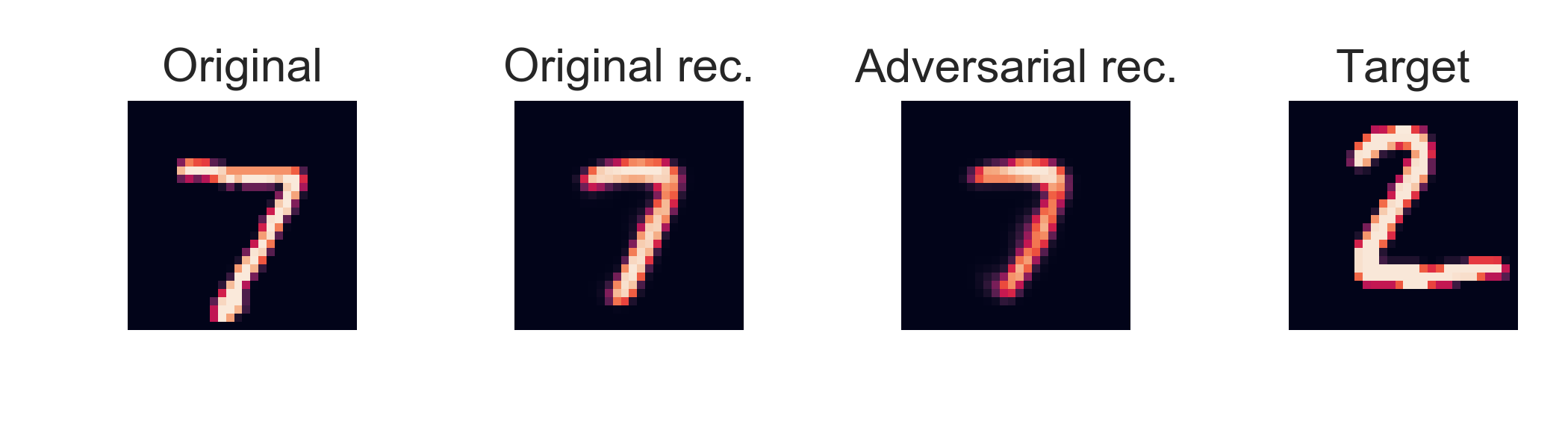}} \\
        \subfloat[Standard VAE, $||\pert||_2 \leq 3$.]{\includegraphics[width=0.49\textwidth]{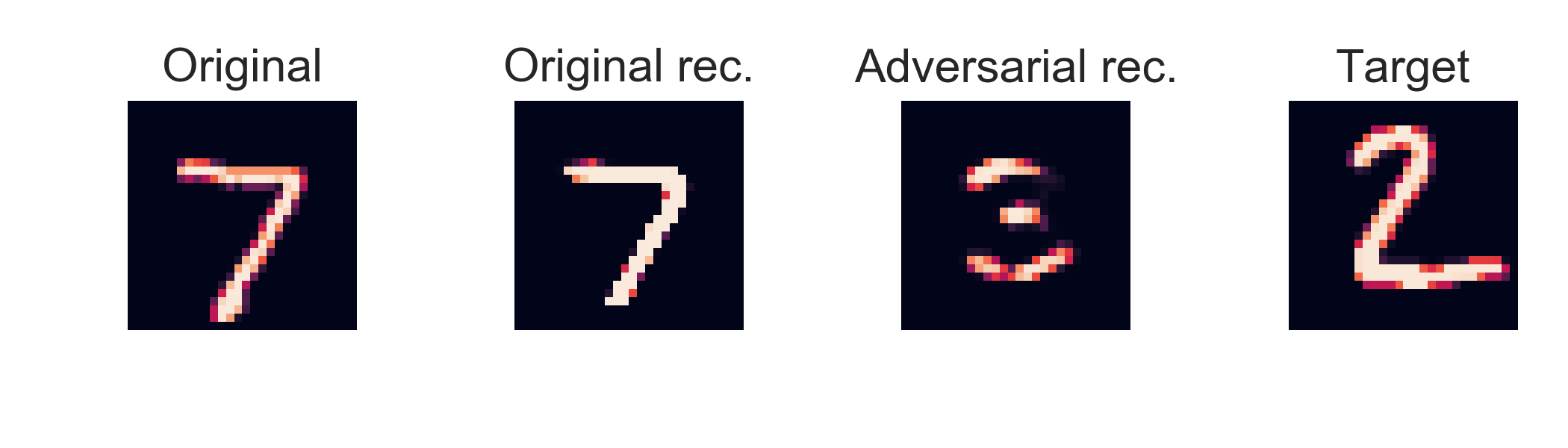}} & \subfloat[Lipschitz-VAE, $||\pert||_2 \leq 3$.]{\includegraphics[width=0.49\textwidth]{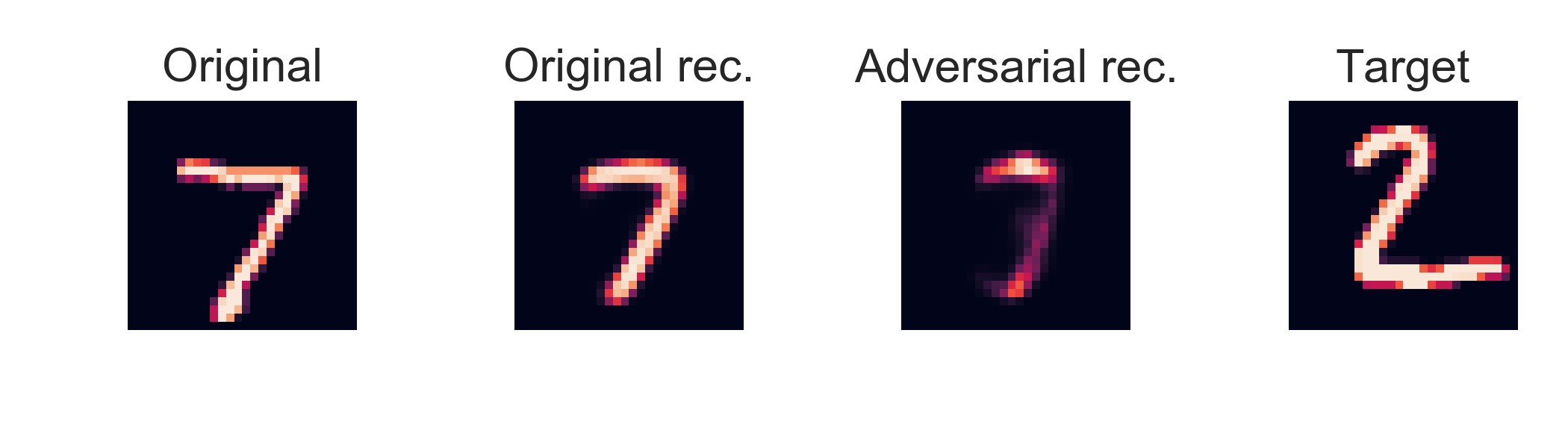}} \\
        \subfloat[Standard VAE, $||\pert||_2 \leq 5$.]{\includegraphics[width=0.49\textwidth]{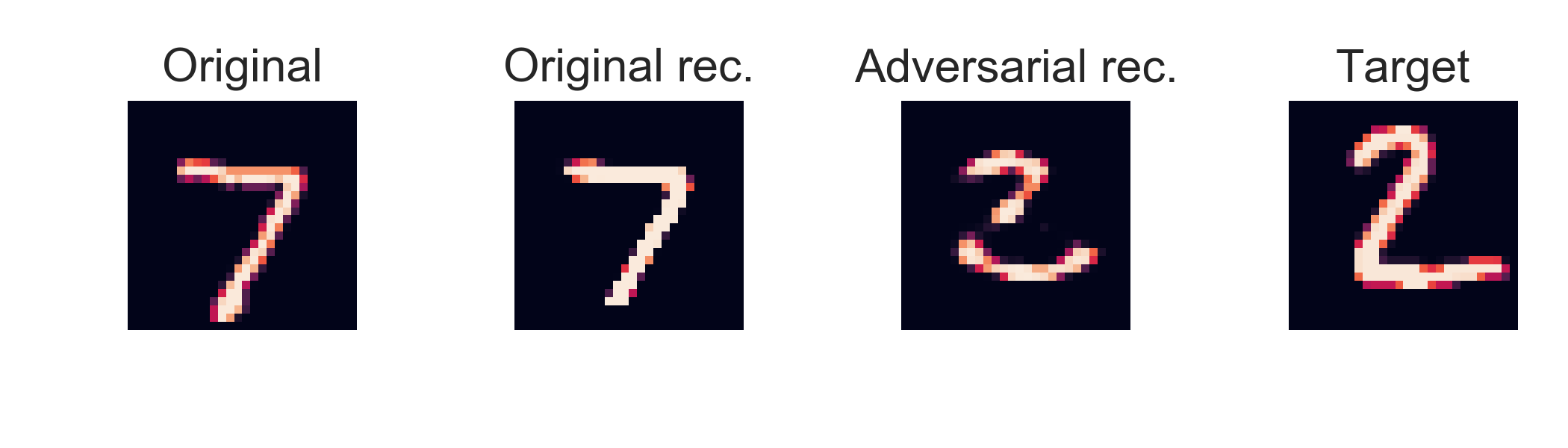}} & \subfloat[Lipschitz-VAE, $||\pert||_2 \leq 5$.]{\includegraphics[width=0.49\textwidth]{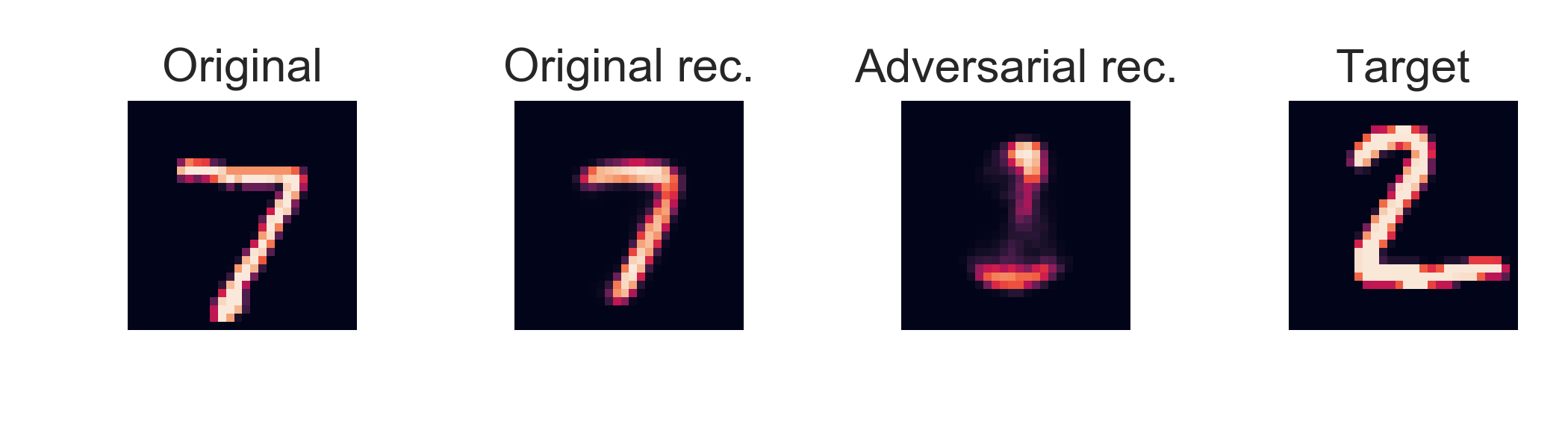}}
    \end{tabular}
    \caption[Latent space attacks on Lipschitz and standard VAEs.]{Representative results from latent space attacks as in \eqref{eq: latent_space_attack_2} on a standard VAE and a Lipschitz-VAE with Lipschitz constants $5$. Each latent space attack looks for an input perturbation $\pert$ such that, applied to an image of a written $7$, the attacked VAE reconstructs an image resembling a written $2$. From left to right in each subfigure: the original image of the written $7$; a reconstruction of the original image, absent input perturbation; a reconstruction of the original image under input perturbation; the target image for the latent space attack, a written $2$. A latent space attack is more successful when reconstructions of the original image under input perturbation more closely resemble the target image. We see latent space attacks are more successful in both the standard and Lipschitz-VAE as the norm of the perturbation $||\pert||_2$ is allowed to increase (moving from top to bottom), but for a given perturbation norm are less successful on the Lipschitz-VAE (right column) than on the standard VAE (left column).
    }
    \label{fig: latent_space_attacks_sidebyside}
\end{figure*}

\section{EXPERIMENTAL SETUP}
\label{app: net_arch}

To properly handle reconstructions on $[0, 1]$-valued data, we let the likelihood in the VAE objective be Continuous Bernoulli  \citep{loaizaganem2019continuous}.

In the Lipschitz-VAEs we train, all activation functions bar the final-layer activations are the GroupSort activation (recall Section \ref{sec: activation_functions}), while in the standard VAEs we train, these are the ReLU. In both types of VAE, the final-layer activation in the encoder standard deviation $\sigma_\phi(\cdot)$ is the sigmoid function to ensure positivity, while the final-layer activation in the deterministic component of the decoder is the sigmoid function to ensure reconstructions are appropriate for binary data. The final layer of the encoder mean takes no activation function.

All models were trained on a 13-inch Macbook Pro from 2017 with 8GB of RAM and 2 CPUs.
\end{appendices}

\end{document}